%% file: main.tex
\documentclass[times, review, 10pt]{elsarticle}
\graphicspath{../}
\usepackage[english]{babel}
\usepackage[utf8x]{inputenc}
\usepackage{listings}
\usepackage{color}
\usepackage{algorithm}
\usepackage{algpseudocode}
\usepackage{verbatim}
\usepackage{soul} % for striking
\usepackage[toc,page]{appendix}
\usepackage{amsfonts}
\usepackage[colorlinks = true,linkcolor =red, urlcolor = blue]{hyperref}
\usepackage{caption}

\usepackage{subcaption}
\usepackage{url}
\usepackage{makecell}
\usepackage{chngcntr}
\usepackage{apptools}
\usepackage{ntheorem}
\usepackage{multirow}
\usepackage{tabularx}
\usepackage{xcolor}
\definecolor{darkred}{RGB}{139, 0, 0}

\usepackage{float}
\usepackage{shuffle}
\usepackage{graphicx}
\usepackage{diagbox}
\usepackage{amsmath} % for \text{} and other math commands
\usepackage{float} % Add this package for more float control
\usepackage{cleveref}
\usepackage{natbib}
\usepackage{booktabs}
\usepackage{multirow}
\usepackage{array}
\usepackage{diagbox}
\usepackage[toc,page]{appendix}

\AtAppendix{\renewtheorem{theorem}{Theorem A.}}

\newcommand{\tri}[1]{{\left\vert\kern-0.25ex\left\vert\kern-0.25ex\left\vert #1 
    \right\vert\kern-0.25ex\right\vert\kern-0.25ex\right\vert}}

\usepackage{eqparbox}

\usepackage[colorinlistoftodos]{todonotes}
\usepackage{booktabs}
\usepackage{lineno}

\journal{Information Science}

\newsavebox{\measurebox}

\begin{document}

\begin{frontmatter}

\title{NODE-AdvGAN:  Improving the transferability and perceptual similarity of adversarial examples by dynamic-system-driven adversarial generative model}

\author[clyde]{Xinheng Xie}
\author[clyde]{Yue Wu}
\author[uga,ok,*]{Cuiyu He}

\affiliation[clyde]{organization={Department of Mathematics and Statistics, University of Strathclyde},%Department and Organization
            addressline={26 Richmond St}, 
            city={Glasgow},
            postcode={G1 1XQ}, 
            country={UK}}

\affiliation[uga]{organization={Department of Mathematics, University of Georgia},%Department and Organization
            addressline={1023 D. W. Brooks Drive}, 
            city={Athens, GA},
            postcode={30605}, 
            country={USA}}
            
\affiliation[ok]{organization={Department of Mathematics, Oklahoma State University},%Department and Organization
            addressline={401}, 
            city={Stillwater, OK},
            postcode={74078}, 
            country={USA}}

\affiliation[*]{Corresponding author: cuiyu.he@uga.edu}
\begin{abstract}
{Understanding adversarial examples is crucial for improving model robustness, as they introduce imperceptible perturbations to deceive models. Effective adversarial examples, therefore, offer the potential to train more robust models by eliminating model singularities. We propose NODE-AdvGAN, a novel approach that treats adversarial generation as a continuous process and employs a Neural Ordinary Differential Equation (NODE) to simulate generator dynamics. By mimicking the iterative nature of traditional gradient-based methods, NODE-AdvGAN generates smoother and more precise perturbations that preserve high perceptual similarity when added to benign images. We also propose a new training strategy, NODE-AdvGAN-T, which enhances transferability in black-box attacks by tuning the noise parameters during training. Experiments demonstrate that NODE-AdvGAN and NODE-AdvGAN-T generate more effective adversarial examples that achieve higher attack success rates while preserving better perceptual quality than baseline models.
}
\end{abstract}

\end{frontmatter}

\input{intro_v3}

\input{Preliminaries}

\input{sec2_NODE}

\input{sec_experimental_results}
\input{conclusion}

\begin{appendices}
\section{Formula: Gradient-based Methods}
\label{app: gradient}
Take a stepsize $0<\alpha\ll 1$. 

\begin{itemize}
    \item {Iterative FGSM (I-FGSM)}:
    \begin{equation}
    x_{n+1} = x_n + \alpha \cdot \text{sign}\big(\nabla_{x} J(\Theta, x_n, y)\big)
     \label{equ: IFGSM} \text{ with }x_0=x,
\end{equation}
where function $\text{sign}: \mathbb{R}\mapsto \{\pm 1, 0\}$, mapping positive numbers to $1$, negative numbers to $-1$ and zero to $0$.
    \item {Momentum Iterative FGSM (MI-FGSM)}:
\begin{equation}
    x_{n+1} = x_n + \alpha \cdot \text{sign}(g^{\text{MI}}_{n+1}) \text{ with }x_0=x
     \label{equ: MIFGSM},
\end{equation}
where 
\[
    g_{n+1}^{\text{MI}} = \mu \cdot g_n^{\text{MI}} + \frac{\nabla_{x} J(\Theta, x_n, y)}{\|\nabla_{x} J(\Theta, x_n, y)\|_1} \text{ subject to }g_0^{\text{MI}}=0,
\]
with \(\mu \geq 0\) being the decay factor for the momentum term.
    \item {Nesterov Iterative FGSM (NI-FGSM)}:
\begin{equation}
    x_{n+1} = x_n + \alpha \cdot \text{sign}(g^{\text{NI}}_{n+1}) \text{ with }x_0=x
     \label{equ: NIFGSM},
\end{equation}
where
\[
g_{n+1}^{\text{NI}} = \mu \cdot g_n^{\text{NI}} + \frac{\nabla_{x} J(\Theta, x_n + \alpha \cdot \mu \cdot g_n^{\text{NI}}, y)}{\|\nabla_{x} J(\Theta, x_n + \alpha \cdot \mu \cdot g_n^{\text{NI}}, y)\|_1} \text{ with } g_0^{\text{NI}}=0.
\]
\end{itemize}

\section{Architecture of the Vector Field}
\label{sec: Architecture of the Vector Field}
{
The architecture of the vector field $\mathcal{N}^{\text{vec}}_{\boldsymbol{\Theta}}$ used in the NODE-AdvGAN model for CIFAR-10 and Fashion-MNIST datasets is detailed in \Cref{tab: Architecture of vector field}. The table outlines the output shape and functionality of each layer, with a particular focus on convolutional layers that utilize varying dilation factors to capture multi-scale features. As mentioned earlier, $(C, H, W)$ represents the input's number of channels, height, and width, respectively. For CIFAR-10, the input size is $(3, 32, 32)$, while for Fashion-MNIST, since the images are resized to $32 \times 32$, the input size is $(1, 32, 32)$.

\begin{table}[]
\centering
\caption{Architecture of the vector field $\mathcal{N}^{\text{vec}}_{\boldsymbol{\Theta}}$ used in the NODE-AdvGAN model on CIFAR-10 and Fashion-MNIST datasets. }
\label{tab: Architecture of vector field}
\begin{tabular}{@{}l|l|p{6cm}@{}} % 使用 p{5cm} 来设置描述列的宽度为 5cm
\hline
\textbf{Layer}   & \textbf{Output shape} & \textbf{Description}                                          \\ \hline
Input            & (C, H, W)               & Input with size (C, H, W)                                     \\
Adding time channel & (C+1, H, W)             & An additional channel is added to represent the time variable \\
Conv + BN + ReLU & (32, H, W)              & Kernel size $= 3 \times 3$, dilation factor $= 1$                 \\
Conv + BN + ReLU & (64, H, W)              & Kernel size $= 3 \times 3$, dilation factor $= 3$                 \\
Conv + BN + ReLU & (64, H, W)              & Kernel size $= 3 \times 3$, dilation factor $= 3 $                \\
Conv + BN + ReLU & (64, H, W)              & Kernel size $= 3 \times 3$, dilation factor $= 3 $                \\
Conv + BN + ReLU & (32, H, W)              & Kernel size $= 3 \times 3$, dilation factor $= 3 $                \\
Conv and Output  & (C, H, W)               & Kernel size $= 3 \times 3$, dilation factor $= 1$                 \\ \hline
\end{tabular}
\end{table}
}
\end{appendices}

\section*{Declaration of generative AI in scientific writing.}
During the preparation of this work, the author(s) used ChatGPT to improve the readability and language of the manuscript. After using this tool/service, the author(s) reviewed and edited the content as needed and take(s) full responsibility for the content of the publication.

\section*{Author Contributions}

Xinheng Xie: Conceptualization, Data curation, Formal analysis, Investigation, Methodology, Software, Writing – original draft.  
Cuiyu He: Project administration, Supervision, Validation, Writing – review and editing.  
Yue Wu: Resources, Supervision, Validation, Writing – review and editing.  
All authors have read and approved the final manuscript.

\bibliographystyle{elsarticle-num}
\bibliography{refs}

\end{document}

%% file: intro_v3.tex
\section{Introduction}
\emph{Adversarial examples} are inputs crafted by adding carefully designed \emph{perturbations} to mislead deep neural networks (DNNs), posing significant security threats to machine learning applications \cite{madry2017towards}. These perturbations are typically imperceptible to the human eye, but can cause models to make incorrect predictions with high confidence  \cite{yuan2019adversarial, biggio2018wild, akhtar2018threat}. Adversarial attacks have been widely studied across various domains, including face recognition  \cite{dong2019efficient, nguyen2020adversarial} and autonomous driving  \cite{xiong2021multi, abdel2021security}, where model reliability is crucial.

\begin{figure}[h]
    \centering
    \begin{minipage}{0.25\textwidth}
        \centering
        \includegraphics[width=\textwidth]{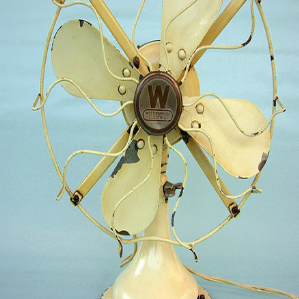} % second image file
    \end{minipage}
    \hspace{2cm} % specify the exact spacing you want between the images
    \begin{minipage}{0.25\textwidth}
        \centering
        \includegraphics[width=\textwidth]{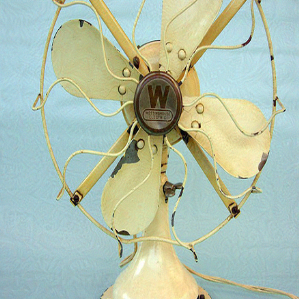} % first image file
    \end{minipage}
    \caption{
    An example of a benign image (left) and its adversarial counterpart (right). 
    }
    \label{Fig: adv example}
\end{figure}

\Cref{Fig: adv example} demonstrates an adversarial example attacking a well-trained Inception-v3  \cite{szegedy2016rethinking} classifier. While the original benign image (left) is correctly classified as an electric fan with $99.7\%$ confidence, the adversarial version (right) is misclassified as an hourglass with $100\%$ confidence. Despite being nearly indistinguishable to the human eye, the adversarial perturbation effectively deceives the model.

The process of generating adversarial images can be mathematically formulated as
\begin{equation}
\label{equ: problem define}
    \boldsymbol{x}_\text{adv} = \boldsymbol{x} + {\delta}(\boldsymbol{x}) \text{ subject to } \|{\delta}(\boldsymbol{x})\|_{p}\leq \epsilon,
\end{equation}
where \(\boldsymbol{x}\) is the benign image, \(\boldsymbol{x}_\text{adv}\) is its adversarial counterpart, and  \({\delta}(\boldsymbol{x})\) denotes the adversarial perturbation. The adversarial algorithm optimizes perturbations to deceive the target model $f$ while adhering to a predefined perturbation budget $\epsilon$, ensuring imperceptibility. We will use $\delta(\boldsymbol{x},f)$ to explicitly denote the need for access to $f$ during inference.

{
In a \emph{white-box attack} scenario where the target model's architecture and parameters are fully accessible, the objective is to find a perturbation $\delta(\boldsymbol{x}, f)$ that maximizes the misclassification likelihood while adhering to a predefined perturbation constraint:
\begin{equation}\label{eqn:optimisation}
    \underset{\|{\delta}(\boldsymbol{x},f)\|_{p}\leq \epsilon}{\text{argmax}}\,
    J({\boldsymbol{\Theta}}, \boldsymbol{x}+{\delta}(\boldsymbol{x},f), \boldsymbol{y}),
\end{equation}
where $J$ is the attack loss function, $\boldsymbol{y}$ is the true label of $\boldsymbol{x}$, and $\boldsymbol{\Theta}$ represents the model parameters. For efficient adversarial generation, \emph{gradient-based} methods are widely used. The {FGSM} \cite{goodfellow2014explaining} crafts perturbations by computing and following the gradient direction of the loss function, while its iterative variant {I-FGSM} \cite{kurakin2018adversarial} significantly enhances both attack effectiveness and visual similarity through carefully controlled multi-step gradient updates within $\epsilon$-constraints. \Cref{Fig: adv example} visualizes this process, demonstrating an adversarial image generated via I-FGSM applied to Inception-v3 \cite{szegedy2016rethinking}.}

{In a \emph{black-box attack}, where the target model’s architecture and parameters are unknown, basic gradient-based methods such as the FGSM \cite{goodfellow2014explaining} and I-FGSM \cite{kurakin2018adversarial} often fail to deceive the model effectively. To address this, \emph{transfer attacks} are commonly employed, which generate adversarial examples using surrogate models to attack the target model. Refined variants, such as Momentum Iterative FGSM (MI-FGSM) \cite{dong2018boosting} and Nesterov Iterative FGSM (NI-FGSM) \cite{dong2019efficient}, enhance transferability through advanced gradient update strategies—momentum-based stabilization and Nesterov-style lookahead, respectively. These approaches adaptively refine perturbations, thereby reducing overfitting to the surrogate and enhancing generalization across diverse model architectures \cite{kurakin2018adversarial}. However, traditional gradient-based methods still rely on hand-crafted iterative gradient directions, which may suboptimally explore the perturbation space and limit transferability.} 

Beyond gradient-based approaches, \emph{machine learning-based} adversarial attack methods leverage generative models or evolutionary algorithms  \cite{xiao2018generating, jandial2019advgan, 9533901, xie2022improving, dai2023advdiff, zhu2024ge, xue2024diffusion}. Among them, Adversarial Generative Adversarial Networks (AdvGAN)  \cite{xiao2018generating} and its variants  \cite{jandial2019advgan, xie2022improving, zhu2024ge} have emerged as powerful alternatives. The generator $\mathcal{G}$ in these methods maps clean inputs to perturbations, i.e., $\boldsymbol{x} \mapsto \mathcal{G}(\boldsymbol{x}) = \delta(\boldsymbol{x})$, while a discriminator ensures perturbations remain imperceptible. The trained generator achieves efficient adversarial example generation through direct inference, eliminating iterative optimization overhead.

While AdvGAN is effective in generating adversarial examples, it struggles to balance visual similarity and adversarial transferability. AdvGAN employs dynamic distillation rather than transfer-attacks for black-box scenarios  \cite{xiao2018generating}, but shows limited transfer efficiency. Moreover, AdvGAN often introduces excessive perturbations, thereby sacrificing visual similarity in favour of higher attack success rates (ASR)  \cite{li2024cgn}. {To address this, we adopt a dynamical systems perspective inspired by the iterative nature of gradient-based attacks, model adversarial example generation as a continuous evolution process. Specifically, unlike iterative gradient-based methods relying on manual updates, we employ Neural Ordinary Differential Equations (NODEs) \cite{chen2018neural} to automatically learn per-step perturbations for adversarial example generation.} NODEs enable dynamic and continuous transformations of input images, allowing finer control over perturbations, which may improve both perceptual similarity and attack robustness. In addition, we propose a new training strategy that significantly improves the transferability of the generated adversarial examples across diverse model architectures. Through the combination of these two techniques, our model demonstrates a significant improvement in generating adversarial examples with higher visual similarity and enhanced ASR in both white-box and black-box settings. 

\subsection*{Our Contribution}
{
This work aims to generate adversarial images with higher ASR, improved visual similarity, and enhanced transferability. These images can be combined with the original training set as an augmented dataset to enhance model robustness. Our key innovation lies in modelling the adversarial example generation process as a continuous evolution, employing a NODE network to generate perturbations, thereby yielding our proposed NODE-AdvGAN model. Additionally, we introduce a novel training strategy to enhance transferability further in our model, referred to as NODE-AdvGAN-T.
}
\paragraph{A dynamic-system perspective} 
The generation of adversarial examples can be interpreted as a continuous-time dynamic process, where an initial benign image is gradually transformed into its adversarial counterpart. {Observing the iterative updates in I-FGSM \eqref{equ: IFGSM}, MI-FGSM \eqref{equ: MIFGSM}, and NI-FGSM \eqref{equ: NIFGSM}, we recognize that some traditional adversarial attack methods implicitly define discrete dynamical systems through perturbation updates.
This process structurally resembles the \emph{numerical approximation} for solving a Ordinary Differential Equation (ODE) with some handcrafted forcing function $F(\cdot)$. Note that such manual designs remain suboptimal for solving the functional extremum problem in Eqn.~\eqref{eqn:optimisation} due to heuristic limitations. Unlike gradient-based methods that rely on {manual selection} of discrete update rules, we }model the perturbation evolution using a NODE framework:

\begin{equation}\label{PDE-model}
 \dot{\boldsymbol{v}}(t) = F(t, \boldsymbol{v},\boldsymbol y, {\boldsymbol{\Theta}_{\textbf{NODE}}}), \quad \boldsymbol{v}(0) = \boldsymbol{x},
 \quad \boldsymbol{v}(T) = \boldsymbol{x}_\text{adv},
\end{equation}
where the function \(F\) represents the learnable vector field guiding the perturbation process, and 
\(\boldsymbol{\Theta}_{\textbf{NODE}}\) denotes the set of trainable parameters in the neural network that parametrizes \(F\). This formulation may offer several advantages:

{\begin{enumerate} 
\item \textbf{Smoothness in the Optimization Process.}
Continuous-time models as in Eqn. (3) produce smooth perturbation trajectories, allowing gradual rather than abrupt updates. This enables finer control over perturbation budget, yielding imperceptible yet effective attacks.

\item \textbf{Adaptively learnable update strategies.}
NODE-AdvGAN can be interpreted as a continuous extension of gradient-based iterative methods. By learning the update function $F$, it avoids the limitations of handcrafted update rules, enabling dynamic adaptation and enhancing the robustness and flexibility of the attack process. 

\item \textbf{Trajectory-Based Generalization.}
Traditional GAN-based attacks typically learn static perturbation mappings, which can overfit the surrogate model. NODE-AdvGAN learns perturbations as dynamic trajectories, capturing broader attack patterns and improving transferability across different architectures.
\end{enumerate}
}

To implement Eqn.~\eqref{PDE-model}, we replace the generator $\mathcal{G}$ in the original AdvGAN \cite{xiao2018generating} with a NODE network, allowing adversarial examples generation to be modelled as a {continuous-time dynamic process}. While NODEs have been previously applied in the context of adversarial robustness  \cite{kang2021stable, li2022defending}, their application for directly generating adversarial examples has not been extensively explored. Experimental results (Section \ref{sec: Experiments}) demonstrate that NODE-AdvGAN consistently outperforms both gradient-based methods and the original AdvGAN in white-box and transfer ASR while preserving higher input similarity.

\paragraph{A novel training strategy}
\label{sec:training_strategy}

Traditional adversarial training methods typically use a fixed perturbation budget  $\epsilon$ throughout both training and testing. However, this approach can lead to {overfitting to a specific model's decision boundary}, thereby reducing the transferability of adversarial examples to unseen architectures. 

The transferability of adversarial attacks refers to the phenomenon where adversarial images generated for one model can also deceive another, even if the second model has a different architecture or was trained on a different dataset  \cite{szegedy2013intriguing}. This transferability reveals a shared vulnerability among models when faced with adversarial perturbations, with significant implications for robustness studies  \cite{carlini2017towards}, attack scalability  \cite{papernot2017practical}, and security risk assessments  \cite{eykholt2018robust}.

To address this issue, we introduce {NODE-AdvGAN-T}, a dynamic noise tuning strategy that optimizes the training perturbation limit ($\epsilon_{\text{train}}$) based on ASR across multiple target models.
Our approach is motivated by the following key observations:
\begin{itemize}
    \item Generators trained under small \( \epsilon_{\text{train}} \) constraints can produce perturbations that retain efficacy when scaled to larger \( \epsilon \). This phenomenon may arise from the generator's parametric perturbation learning under strict budgets, implicitly capturing attack patterns generalizable beyond $\epsilon_{\text{train}}$.

    \item Compared to larger perturbations, smaller \( \epsilon_{\text{train}} \) may encourage the generator to learn more essential and representative features, resulting in perturbations with better generalization across different models.

    \item Since different target models exhibit varying sensitivities to perturbations, a fixed perturbation budget is unlikely to achieve optimal performance across all models. Therefore, a dynamic perturbation adjustment strategy based on the target model’s attack performance is necessary.
    \end{itemize}

Thus, we tune $\epsilon_{\text{train}}$ dynamically to maximize ASR across different classification architectures while keeping the test perturbation budget $\epsilon$ fixed. Through this approach, NODE-AdvGAN-T {automatically learns an optimal balance between attack strength and imperceptibility}, leading to significantly {improved transfer ASR}. Details of the algorithm are provided in \Cref{alg: generalizability}. {Importantly}, this algorithm is also applicable to the original AdvGAN framework (see Section \ref{ssec: Ablation Experiments}), suggesting its potential adaptability to different neural network-based generators. 

We emphasize that $\epsilon$ is kept fixed during evaluation to ensure consistent comparison with other adversarial attack methods, typically using a fixed perturbation budget. Moreover, in black-box settings where the optimal $\epsilon_{\text{train}}$ cannot be directly determined (due to inaccessible target model decision boundaries), selecting a standardized $\epsilon_{\text{test}}$ becomes necessary for fair evaluation of attack performance.

\paragraph{Organisation} 
The remaining structure of this paper is outlined as follows. \Cref{sec: Preliminaries} reviews the foundational concepts of AdvGAN and NODE. \Cref{sec:secproposed} elaborates on the architecture of the proposed NODE-AdvGAN, detailing the proposed algorithms with loss functions and the training strategy for NODE-AdvGAN-T. {\Cref{sec: Experiments} presents results of ablation studies and comparison with baseline models on both targeted and untargeted attacks on multiple widely used datasets.} Finally, we conclude our study in \Cref{sec: Conclusion}.

%% file: Preliminaries.tex
\section{Preliminaries}
\label{sec: Preliminaries}
\subsection{Adversarial Generative Adversarial Network (AdvGAN)}
{
As powerful tools in the field of generative modelling, Generative Adversarial Networks (GANs)  \cite{goodfellow2020generative} have been extensively applied in various image generation tasks, renowned for their ability to create highly realistic and detailed images  \cite{ledig2017photo, li2022comprehensive, karras2018progressive}. The key to the success of GANs lies in their unique adversarial training framework, consisting of a generator and a discriminator. This framework enables GANs to generate high-quality, synthetic images through an adversarial process, where the generator progressively improves its output to fool the discriminator into accepting the images as real.}

{AdvGAN  \cite{xiao2018generating} adapts the generative adversarial framework specifically for the generation of adversarial examples. In this model, the generator is responsible for producing adversarial examples, while the discriminator's task is to distinguish between these generated examples and real images. Through this process, the generator takes the genuine image as the input and creates noise aiming to mislead the target model and ensures that the noise is subtle enough to remain undetected by the discriminator. Models trained in this manner do not need to access the target  classification model's architecture during the testing phase, enabling more efficient generation of adversarial examples. This attribute allows the attack to operate independently of the target model’s specifics once the adversarial generator has been trained. 
}

\subsection{Neural Ordinary Differential Equations (NODEs)}\label{sec:node}
{
The NODE framework models a continuous function that maps from \(\mathbb{R}^{d}\) to \(\mathbb{R}^{d}\) by employing an ODE as follows: for any given input \(\mathbf{h}_0 \in \mathbb{R}^{d}\), the output of the NODE model is represented by \({h}(T)\) (or, more generally, \({h}(t)_{t \in [0, T]}\)), where the function \({h}(t)\) satisfies the differential equation:
\begin{equation}\label{NODE}
\frac{\mathrm{d}{h}(t)}{\mathrm{d}t}=\mathcal{N}^{\text{vec}}_{\boldsymbol{\Theta}}({h}(t), t), \quad {h}(0) = \mathbf{h}_0.
\end{equation}
In this formulation, the vector field \(\mathcal{N}^{\text{vec}}_{\boldsymbol{\Theta}}\) is parameterized by a neural network with trainable parameters \(\boldsymbol{\Theta}\). This neural network defines the dynamics of {\(h(t)\)} over time, thus enabling the modelling of complex, continuous transformations of the input data. According to Picard's Theorem, the initial value problem \eqref{NODE} has a unique solution provided that \(\mathcal{N}_{\boldsymbol{\Theta}}^{\text{vec}}\) is Lipschitz continuous, which can be ensured by using finite weights and activation functions such as {ReLU} or {Tanh}  \cite{chen2018neural}. }

{
In our model, we apply NODE for the generator.
More specifically, we let \(\mathbf{h}_0\) be the initial clean image \( \boldsymbol{x}\) and \({h}(T)\) as the resulting adversarial example \( \boldsymbol{x}_\text{adv}\). 
}

%% file: sec2_NODE.tex
\section{NODE-AdvGAN for adversarial image}\label{sec:secproposed}
{
\subsection{Problem Statement}
{Consider an image sample \( \boldsymbol{x}\), lying in the feature space \(\mathcal{X}\), which is a subset of the tensor space \(\mathbb{R}^{C \times H \times W}\) ( $\cong\mathbb{R}^{C}\otimes \mathbb{R}^{H}\otimes \mathbb{R}^W $). Here, \(C\), \(H\), and \(W\) represent the number of channels, height, and width of the image, respectively.} Each sample \( \boldsymbol x\) corresponds to a true class label \(\boldsymbol y \in \mathcal{Y}\). Let \(f:\mathcal{X}\to\mathcal{Y}\) denote a classifier that maps input \( \boldsymbol x\) to a predicted classification label \(\boldsymbol y'\), expressed as \(f( \boldsymbol x) = \boldsymbol y'\).  Additionally, define \(f_l\) as the function within the classifier outputting logits, with \(f_l^i( \boldsymbol x)\) representing the logit value for the \(i\)th class. Given an instance \( \boldsymbol{x}\), the objective of an adversary is to create adversarial images \( \boldsymbol x_{\text{adv}}\in \mathbb{R}^{C \times H \times W}\).  {Define the perturbation vector as \({\delta(\boldsymbol{x})} =  \boldsymbol{x}_{\text{adv}} -  \boldsymbol{x}\), where \( \boldsymbol{x}_{\text{adv}}\) is the adversarial image and \( \boldsymbol{x}\) is the original input. The parameter \(\epsilon\) represents the maximum allowable strength of the noise, ensuring \(\|{\delta(\boldsymbol{x})}\|_\infty \leq \epsilon\). This bounds each component of the perturbation \({\delta(\boldsymbol{x})}\) by \(\epsilon\) in absolute value.} In an untargeted attack, the goal is to have \(f( \boldsymbol x_{\text{adv}}) \neq \boldsymbol y\). For a targeted attack, the aim is \(f( \boldsymbol x_{\text{adv}}) = \boldsymbol y_{\text{target}}\), where \(\boldsymbol y_{\text{target}}\) is the specified target class.
}

\subsection{Our Proposed NODE-AdvGAN} \label{sec:proposed}
\begin{figure}
    \centering
    \includegraphics[width=0.9\linewidth]{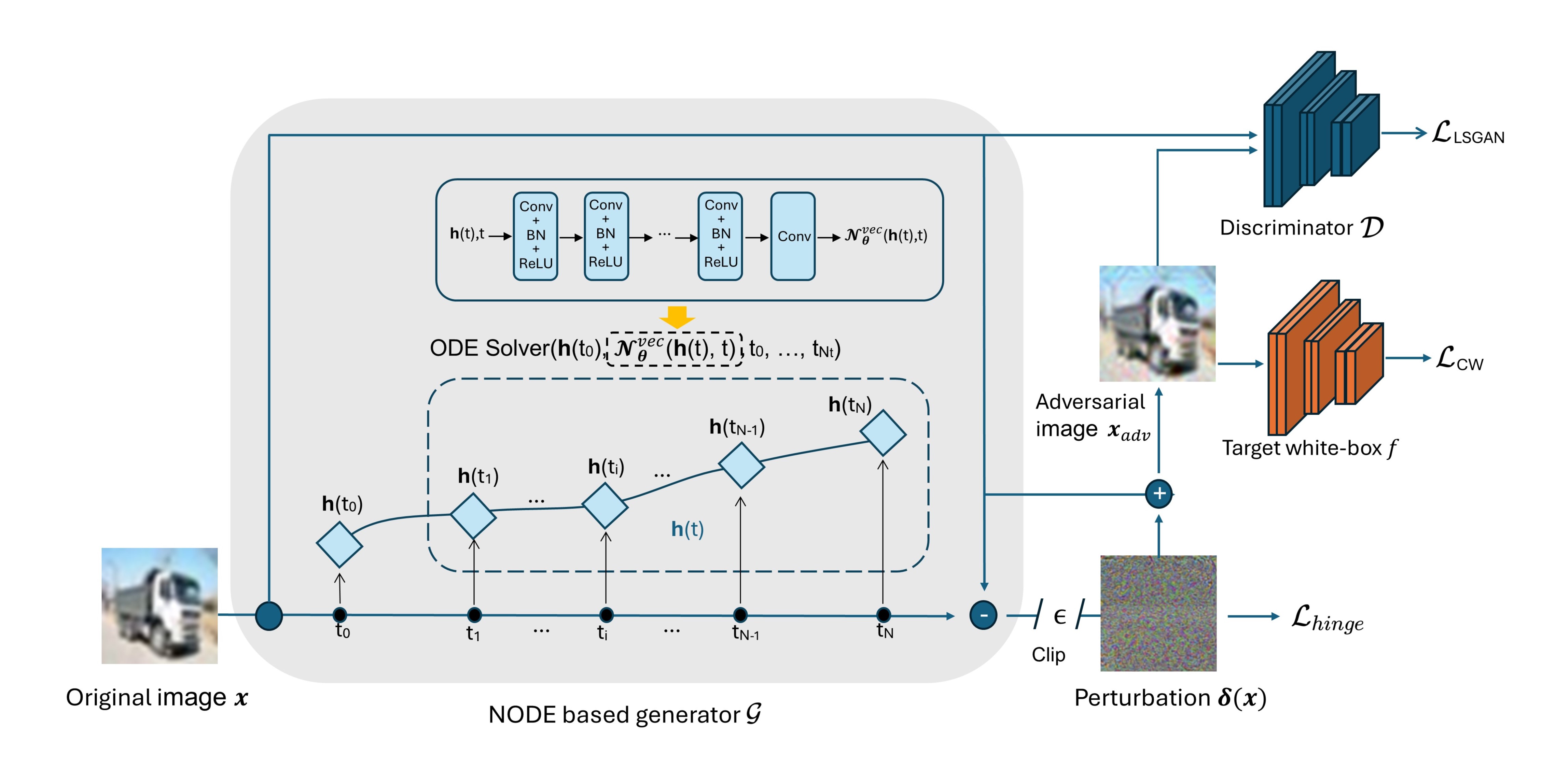}
    \caption{Architecture of the NODE-AdvGAN.}
    \label{fig:NODEAdvGAN framework}
\end{figure}

{
Similar to conventional AdvGAN models, our model consists of a generator \(\mathcal{G}\) and a discriminator \(\mathcal{D}\). For white-box attacks on a target model \(f\), the overall structure of NODE-AdvGAN is presented in \Cref{fig:NODEAdvGAN framework},
} where losses mentioned therein can be found in \cref{ssec: loss function}.

We use the ODE system \eqref{NODE} as the dynamics for generating perturbation noise. 
The vector field \(\mathcal{N}^{\text{vec}}_{\boldsymbol{\Theta}}(\mathbf{h}(t), t)\) is modelled through a CNN-based neural network, which consists of 6 CNN layers. It is worth noting that we have the compatibility with any advanced image-to-image architecture for the vector field \(\mathcal{N}^{\text{vec}}_{\boldsymbol{\Theta}}\).

{
The detailed structure of the 6-layer vector field includes two types of layers: \(\text{Conv} + \text{BN} + \text{ReLU}\) and Conv.  {The term `Conv' refers to a dilated convolutional layer with \(3 \times 3\) filters,  `BN' denotes batch normalization, and `ReLU' denotes rectified linear units.} Layers 1 to 5 follow the \(\text{Conv} + \text{BN} + \text{ReLU}\) structure, and the 6th layer is a Conv layer. Each layer's channel count also shown in the \Cref{tab: Architecture of vector field}  in Appendix \ref{sec: Architecture of the Vector Field}, with the image size remaining constant throughout. The first layer has \(c + 1\) input channels, where \(c\) represents the number of image channels and the additional channel corresponds to the time variable. In the final layer, the number of output channels is set to \(c\).
For the initial and final layers, a dilation factor is set to 1, corresponding to standard convolution without dilation. In contrast, a dilation factor of 3 is applied to convolution layers $2$ through $5$. This design choice ensures that the receptive field size reaches $29$,  {which is nearly the entire height and width ($32\times 32$) of the images in the experimental dataset.}
}

{
Another key feature of our model is its flexibility in adjusting its depth by varying the number of time integration steps, \( N \). In our experiments, we set \( N \) at values between 2 and 8, striking a balance between training time and model performance.  We employ the Euler method for numerical integration and select a final time parameter \( T = 0.05 \).
}

{
For the discriminator, we follow the architectures used in AdvGAN \cite{xiao2018generating}, which are similar to those in the image-to-image translation literature  \cite{isola2017image,zhu2017unpaired}. The discriminator consists of a series of convolutional layers with Leaky ReLU activations and BN for stabilization. 
}

\subsection{Training Strategy of NODE-AdvGAN-T}
\label{ssc:Training Strategy of NODE-AdvGAN-T}
{
To enhance the transferability of adversarial attacks, we propose a novel training strategy for adversarial generative models, particularly during the adversarial image generation phase of AdvGAN. This requires controlling the perturbation magnitude to ensure compliance with a predefined budget \(\epsilon\), which represents the maximum allowable noise across all attack methods. The budget is enforced by truncating perturbations from the generator \(\mathcal{G}\) as follows:
\[
\delta(\boldsymbol{x}) = \text{Clip}(\mathcal{G}(\boldsymbol{x}), -\epsilon, \epsilon).
\]
Here, the \(\text{Clip}\) function limits every element in vector \(\mathcal{G}(\boldsymbol{x})\) to a specified range 
\([-\epsilon, \epsilon]\), mathematically equivalent to:
\[ \text{Clip}(\mathcal{G}_i(\boldsymbol{x}), -\epsilon, \epsilon) = \min\left(\max\left(\mathcal{G}_i(\boldsymbol{x}), -\epsilon\right), \epsilon\right) \]

where $\mathcal{G}_i(\boldsymbol{x})$ represent the i-th element of the vector $\mathcal{G}(\boldsymbol{x})$.
Traditionally, the noise level, denoted by \( \epsilon_{\text{train}} \), used in the training phase is set equal to \( \epsilon \) in the testing phase. In our approach, however, we treat \( \epsilon_{\text{train}} \) as a tunable hyperparameter in the training phase, while \( \epsilon \) in the testing phase remains a fixed, user-defined value. Consequently, \( \epsilon_{\text{train}} \neq \epsilon \), as detailed in \Cref{alg: generalizability}. Experimental validation suggests that this strategy significantly enhances transferability over traditional methods.

\setlength{\textfloatsep}{10pt} % Reduce space above and below floats
\setlength{\floatsep}{10pt} % Reduce space between floats

\begin{algorithm}[!ht]
\caption{
NODE-AdvGAN-T for Improved Attack Transferability
}
\label{alg: generalizability}
\begin{algorithmic}[1]
\small
\Require
\Statex Training set \(X_{\text{train}}\) and testing set \(X_{\text{test}}\), target classifier \(f(\cdot)\), generator \(\mathcal{G}(\cdot)\), discriminator \(\mathcal{D}(\cdot)\),   \(\epsilon_{\text{train}}\) and \(\epsilon\).
\State\textbf{Training Phase:}
\State Initialize weights of $\mathcal{G}(\cdot)$ and $\mathcal{D}(\cdot)$; set epoch number $M$
\For{$i = 1$ \textbf{to} $M$}
    \For{\textbf{each} batch $\{ \boldsymbol{x}\}$ from $X_{\text{train}}$}
        \State Generate perturbations within training intensity limit: 
        \[\delta_\text{train}(\boldsymbol{x}) = \text{Clip}(\mathcal{G}(\boldsymbol{x}), -\epsilon_\text{train}, \epsilon_\text{train})\]
        \State Apply perturbations and ensure valid pixel range: 
\[\boldsymbol{x}_{\text{adv}} = \text{Clip}(\boldsymbol{x} + \delta_\text{train}(\boldsymbol{x}), 0, 1)\]
        \State Compute loss functions for $\mathcal{D}(\cdot)$ and $\mathcal{G}(\cdot)$ using $\boldsymbol{x}_{\text{adv}}$ and update weights via backpropagation
    \EndFor
\EndFor

\State \textbf{Testing Phase:}
\For{\textbf{each} batch $\{ \boldsymbol{x}\}$ from $X_{\text{test}}$}
    \State Generate adversarial images: 
        \[\boldsymbol{x}_{\text{adv}} = \text{Clip}\left(\boldsymbol{x} + \text{Clip}(\mathcal{G}(\boldsymbol{x}), -\epsilon, \epsilon ), 0, 1 \right)\]
\EndFor
\end{algorithmic}
\end{algorithm}

{
We define the model that achieves optimal white-box attack performance when \(\epsilon = \epsilon_{\text{train}}\) as NODE-AdvGAN, and the model that employs the best \(\epsilon_{\text{train}}\) after tuning as NODE-AdvGAN-T. In this paper, we identify the optimal \(\epsilon_{\text{train}}\) by performing hyperparameter tuning ablation experiments specifically on the VGG16 model. While tuning \(\epsilon_{\text{train}}\) for each model configuration can further improve transferability, our approach demonstrates that selecting an near-optimal \(\epsilon_{\text{train}}\) for one model can generalize effectively across others. Although this selected \(\epsilon_{\text{train}}\) may not be the absolute best in all cases, NODE-AdvGAN-T consistently shows significant improvements in transferability. Future research will investigate more efficient methods for determining the optimal \(\epsilon_{\text{train}}\) to enhance robustness across diverse models and settings further.
}

\subsection{Loss Function}
\label{ssec: loss function}
{
We follow the loss definitions from AdvGAN  \cite{xiao2018generating}, where the loss functions for the generator and discriminator, denoted as \(\mathcal{L}_G\) and \(\mathcal{L}_D\), are minimized as  \(\underset{\mathcal{G}}{\min} \, \mathcal{L}_{G}(\mathcal{G}, \mathcal{D})\) and \(\underset{\mathcal{D}}{\min} \, \mathcal{L}_{D}(\mathcal{G}, \mathcal{D})\) in the training phase. Our notation differs from  \cite{xiao2018generating} to emphasise the specific loss configuration used in their experiments.

More precisely, the generator aims to minimize the following loss function, which comprises three terms:
{
\begin{equation}
\label{equ: total loss}
\mathcal{L}_G(\mathcal{G},\mathcal{D}) = \mathcal{L}_{\text{CW}}(\mathcal{G}) + \alpha\mathcal{L}_{LSGAN_G}(\mathcal{G},\mathcal{D}) + \beta\mathcal{L}_{hinge}(\mathcal{G}).
\end{equation}
}

{
The term $\mathcal{L}_{\text{CW}}$, designed to deceive the target model $f$, is based on the Carlini-Wagner (CW) loss  \cite{carlini2017towards}. Recall that $f_l({ \boldsymbol{x}})$ represents the  logits (pre-softmax outputs) of the output of target model $f$ given the input ${ \boldsymbol{x}}$ and
the subscript $i$  in  \(f_l^i(\cdot)\)
represents the logits for the $i^{th}$ category. 
}
\begin{itemize}
    \item For targeted attacks, it is defined as: {
\[
    \mathcal{L}_{\text{CW}}(\mathcal{G}) = \mathbb{E}_{\boldsymbol{x}} \left[\max\left(\max_{i \neq target}\{f_l^i({\boldsymbol{x}} + \mathcal{G}(\boldsymbol{x}))\} - f_l^{target}(\boldsymbol{x} + \mathcal{G}(\boldsymbol{x})), \kappa \right)\right],
\]

where \(\mathbb{E}_{\boldsymbol{x}}\) represents the expectation over all data points \(\boldsymbol{x}\) i.e. \(\mathbb{E}_{\boldsymbol{x}} \equiv \mathbb{E}_{\boldsymbol{x} \sim {\text{data}}}\) \footnote{In practice, we use a batch of data to compute this expectation.} , the constant \(\kappa \ge 0 \)  is a confidence parameter used to control the adversarial strength, and $target$ is the target class index to fool $f$ on ${\boldsymbol{x}} + \mathcal{G}(\boldsymbol{x})$.
}
    \item For untargeted attacks, the $\mathcal{L}_\text{adv}$ is expressed as:
    \[
    \mathcal{L}_{\text{CW}}(\mathcal{G}) = 
    \mathbb{E}_{\boldsymbol{x}}\left[ \max \left( \max_{i \neq true}\{f_l^{true}(\boldsymbol{x} + \mathcal{G}(\boldsymbol{x})) - f_l^i(\boldsymbol{x} + \mathcal{G}(\boldsymbol{x})) \}, \kappa \right) \right].
\]
Here, \(true\) denotes the ground-truth class of $\boldsymbol{x}$, and the loss maximizes the gap between the true class logit and the highest competing logit.
\end{itemize}

To bound the magnitude of the perturbation noise $\mathcal{G}({\boldsymbol{x}})$, a soft hinge/penalty loss on the {Euclidean  $L_2$ norm} is added as:
\begin{equation}\label{hinge}
\mathcal{L}_{hinge}(\mathcal{G}) = \mathbb{E}_{\boldsymbol{x}} \max(0, \| \mathcal{G}(\boldsymbol{x})\|_2 - c),
\end{equation}
where \(c \ge 0\) denotes a user-specified hyper-parameter.
The model will start penalize  the $L^2$ norm of the noise $\mathcal{G}(\boldsymbol{x})$ when its scale exceeds $c$.
This term also helps stabilize the training of the GAN  \cite{isola2017image}. 

{
The least squares objective function LSGAN  \cite{mao2017least}is defined as:
\[
\mathcal{L}_{LSGAN_G}(\mathcal{G},\mathcal{D}) = \frac{1}{2} \mathbb{E}_{\boldsymbol{x}}[(\mathcal{D}(\boldsymbol{x} + \mathcal{G}(\boldsymbol{x})) - 1)^2].
\]
For the discriminator \(\mathcal{D}(\cdot)\), the range of 
$
\mathcal{D}(\cdot) \in [0, 1],
$
 where a value of $\mathcal{D}(\cdot)$ closer to 1 indicates that the discriminator considers the image to be more realistic, and vice versa. Minimizing this loss function helps the generator produce adversarial images $\boldsymbol{x} + \mathcal{G}(\boldsymbol{x})$ closer to the original as much as possible and thus makes it more difficult to detect or discern. 

For the discriminator, training minimizes the following loss function:

\[
\mathcal{L}_D(\mathcal{G}, \mathcal{D}) = \frac{1}{2}\mathbb{E}_{\boldsymbol{x}}[(\mathcal{D}(\boldsymbol{x}) - 1)^2] + \frac{1}{2}\mathbb{E}_{\boldsymbol{x}}[\mathcal{D}(\boldsymbol{x} + \mathcal{G}(\boldsymbol{x}))^2],
\]

where \(\mathcal{L}_D(\mathcal{D}, \mathcal{G})\) uses the least squares objective from LSGAN  \cite{mao2017least}. Minimizing \(\mathcal{L}_D\) trains the discriminator to recognize the original image \(\boldsymbol{x}\) as authentic and the perturbed adversarial image \(\boldsymbol{x} + \mathcal{G}(\boldsymbol{x})\) as altered. The discriminator thus assesses the similarity between adversarial and original images, encouraging the generator to produce more convincing adversarial examples.
}

%% file: sec_experimental_results.tex
\section{Experiments}
\label{sec: Experiments}
{This section presents the experimental results of our NODE-AdvGAN and NODE-AdvGAN-T models, along with comparisons to gradient-based models and the original AdvGAN  model. The datasets and experimental settings are documented in \Cref{ssec: Environment Setup}. In \Cref{ssec: Ablation Experiments}, we perform ablation studies and parameter selection using untargeted attacks on the CIFAR-10 dataset. \Cref{ssec: Results and Comparisons} contains our main experiments, where we test adversarial examples across different datasets and present results for white-box and transferability attacks to evaluate the robustness of the models. Specifically, \Cref{sssec: Results on Untargeted Attack} covers untargeted attacks, while \Cref{sssec: Results on Targeted Attack} focuses on targeted attacks. The codes are available at GitHub\footnote{\url{https://github.com/xiexinheng/NODE-AdvGAN}}.}
\subsection{Environment Setup}
\label{ssec: Environment Setup}
\subsection*{Datasets}
{In our study, we choose the Fashion-MNIST (FMNIST)  \cite{xiao2017fashion} and CIFAR-10  \cite{krizhevsky2009learning} datasets as benchmark datasets to evaluate our method's performance. The FMNIST dataset comprises single-channel greyscale images structured into ten different classes. It includes a training set of 60,000 images, with 6,000 images per class, and a test set of 10,000 images, with 1,000 images per class. Originally, each image is $1 \times 28 \times 28$ pixels in size. In the experiments, images are resized to a resolution of $1 \times 32 \times 32$ pixels. The CIFAR-10 dataset features three-channel colour images, similarly divided into ten categories. It comprises a training set of 50,000 images and a test set of 10,000 images, with each category represented by 5,000 and 1,000 images, respectively. Each image in this dataset is $3 \times 32 \times 32$ pixels.}

\subsection*{Target Image Classification Models}\label{ssec:models}
{
In selecting target models, we choose three commonly used classification architectures: the Visual Geometry Group Network (VGG)  \cite{simonyan2014very}, Residual Network (ResNet)  \cite{he2016deep}, and Densely Connected Convolutional Network (DenseNet)  \cite{huang2017densely}.
For each architecture, we select two configurations: VGG16 and VGG19, ResNet18 and ResNet34, DenseNet121 and DenseNet169. The classification performance of these architectures on clean samples is shown in \Cref{tab: models_acc}.\footnote{Pretrained weights are available at GitHub} To improve robustness and accuracy, we apply random cropping with padding, followed by normalization.
}

\begin{table}[htbp]
\centering\resizebox{0.95\textwidth}{!}{%
\begin{tabular}{@{}ccccccc@{}}
\toprule
\textbf{}  & \textbf{VGG16} & \textbf{VGG19} & \textbf{ResNet18} & \textbf{ResNet34} & \textbf{DenseNet121} & \textbf{DenseNet169} \\ \midrule
\textbf{FMNIST}   & 95.13\%        & 94.87\%        & 94.76\%           & 94.70\%           & 94.90\%              & 94.56\%              \\
\textbf{CIFAR-10} & 94.00\%        & 94.26\%        & 93.64\%           & 94.16\%           & 94.29\%              & 94.27\%              \\ \bottomrule
\end{tabular}%
}
\caption{Accuracy (\%) of  image classification models.}
\label{tab: models_acc}
\end{table}

\subsection*{Baseline Methods}
\label{ssec:Baseline methods}
{
To validate the effectiveness of our method, we compare it with classical gradient-based methods, including FGSM  \cite{goodfellow2014explaining}, I-FGSM  \cite{kurakin2016adversarial}, MI-FGSM  \cite{dong2018boosting}, and NI-FGSM  \cite{dong2019efficient}, as mentioned in the introduction section. We also compare with the original AdvGAN  \cite{xiao2018generating}, which employs a different generator. Since our approach differs from the original AdvGAN only in the generator component, this comparison can also be considered an ablation experiment. For both the FMNIST and CIFAR-10 datasets, we set the user-defined maximal perturbation budget $\epsilon = 15/255$ for all methods. For the iterative methods I-FGSM, MI-FGSM, and NI-FGSM, we set the step size $\alpha = 2/255$ and the number of iterations to 10, with a decay factor $\mu = 1$ specifically for MI-FGSM and NI-FGSM. For the training of both the original AdvGAN and our method, the datasets undergo the same data augmentation techniques, specifically random cropping with padding, as used in training classification models.
}

\subsection*{Evaluation Metrics} 

To evaluate method effectiveness, we use Attack Success Rate (ASR) and measure image similarity with Peak Signal-to-Noise Ratio (PSNR)  \cite{gonzalez2009digital} and Structural Similarity Index (SSIM)  \cite{wang2004image}. SSIM and PSNR calculations are performed using the piqa Python package (version 1.3.2)  \cite{piqa}. Our goal is to maximize the ASR while ensuring high values for PSNR and SSIM to maintain image quality.

\subsection*{Other Experimental Settings}
\label{sec:Expaermental setting}
{
The batch size is set to $256$ during the training, and the training spans $150$ epochs. The initial learning rate is set at $0.002$ and is reduced by half every 60 epochs. In terms of loss settings, we set \(\kappa = 0\), \(c = 0.1\), \(\alpha = 0.01\), and \(\beta = 0.01\) in \cref{equ: total loss}; for NODE configurations, we set \(T = 0.05\) and time integration steps \(N = 5\).
}

{
Our experimental setup employs the PyTorch framework   \cite{NEURIPS2019_9015} for model training. The experiments are executed on a server configured with the Linux-$5.4.0$ operating system and a Python $3.9$ runtime environment. This server boasts $1.4$ TiB of memory and utilizes an Intel Xeon(R) Gold 6240 CPU, operating at $2.60$GHz with $72$ cores. It features an NVIDIA GeForce RTX 3080 Ti GPU with 12GB GDDR6X memory. To implement NODEs, we leverage version $0.2.3$ of the torchdiffeq package, which is optimized explicitly for GPU computations and facilitates reduced memory usage during backpropagation.
}

\subsection{Parameter Selection and Ablation Experiments}
\label{ssec: Ablation Experiments}

In this subsection, we conduct comprehensive ablation studies to evaluate the impact of key components and optimize critical parameters of our model. Specifically, we analyze the effects of $\mathcal{L}_{\text{CW}}$, $\mathcal{L}_{\text{LSGAN}_G}$, and $\mathcal{L}_{\text{hinge}}$ on model performance, while optimizing the values of $\alpha$ and $\beta$. Additionally, we examine the effect of the NODE system and its time integration step \(N\), identifying the optimal \(N\). Finally, we evaluate the effectiveness of our proposed training strategy for adjusting $\epsilon_{\text{train}}$ and selecting a relatively optimal $\epsilon_{\text{train}}$. All experiments are performed using white-box untargeted attacks on the CIFAR-10 dataset.

\begin{figure}[htbp]
    \centering
    % \begin{minipage}{0.7\textwidth}
    %     \centering
    \includegraphics[width=0.7\linewidth]{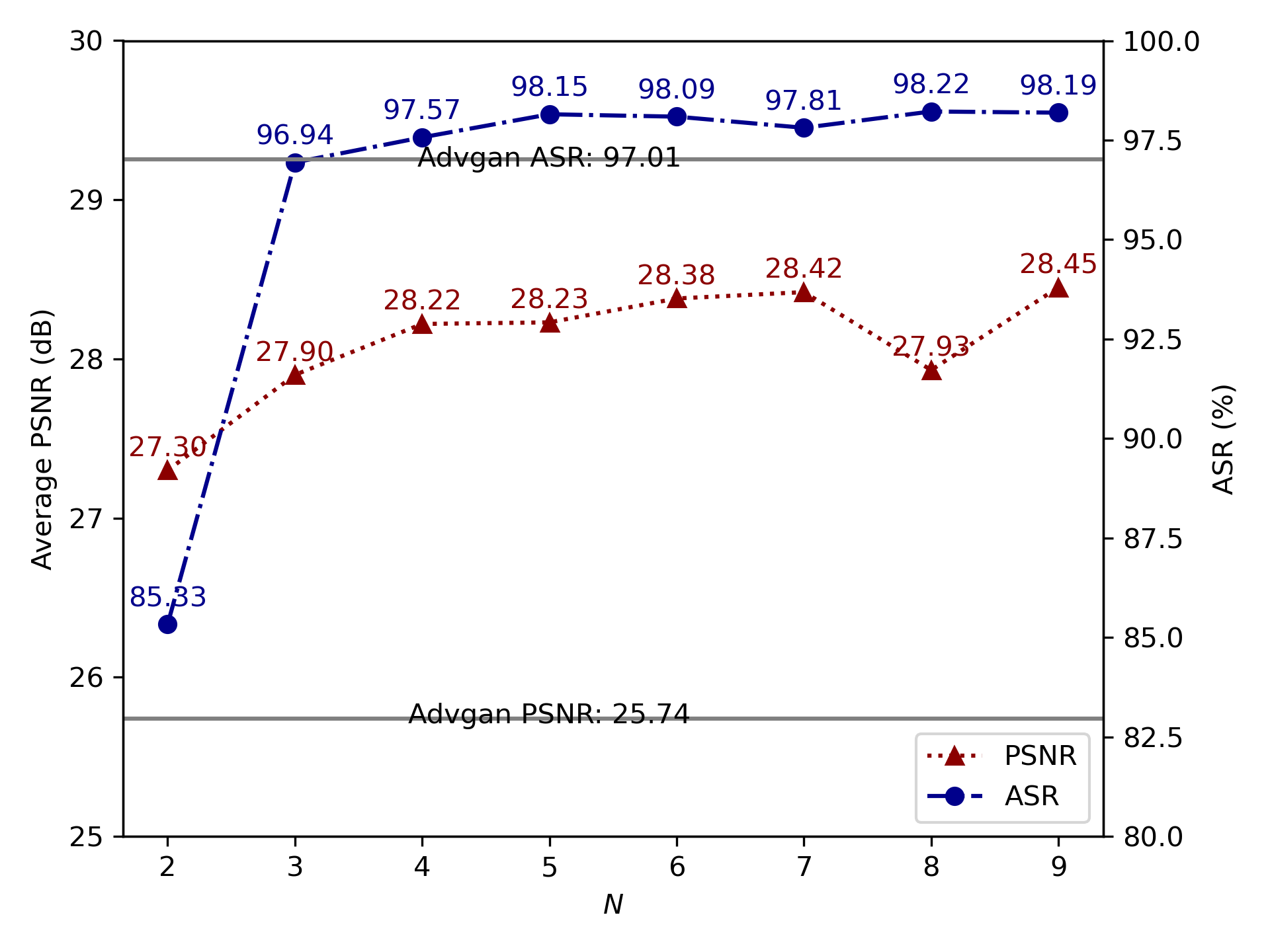}
    \caption{ASR (Blue) and PSNR (Red)  for various \(N\) values in NODE-AdvGAN model compared to baseline AdvGAN.}
    \label{fig: N_t plot}
\end{figure}

{Training generators to create adversarial examples requires balancing the weights of the three loss components. To achieve an optimal balance, we test different values of $\alpha$ and $\beta$ (1, 0.1, 0.01, 0.001, and 0.0001) and evaluate performance using ASR and SSIM, selecting the final parameters based on the average of these scores.} Using NODE-AdvGAN, we conduct white-box attacks on the VGG16 model with the CIFAR-10 dataset, setting a perturbation limit {\(\epsilon_{train} =\epsilon\) of \(15/255\).} The results for ASR, SSIM, and their average are shown in the heatmap in \Cref{fig: score heatmap plot}.
We find that increasing the weight of $\mathcal{L}_{\text{CW}}$ leads to higher ASR.
 However, without the constraints imposed by $\mathcal{L}_{\text{LSGAN}_G}$ and $\mathcal{L}_{\text{hinge}}$, the overall performance drops. {Meanwhile, $\mathcal{L}_{\text{LSGAN}_G}$ and $\mathcal{L}_{\text{hinge}}$ serve to constrain the optimization process and enhance similarity, fulfilling their intended roles as designed.} {Based on these results, we set \(\alpha = 0.01\) and \(\beta = 0.01\) as the optimal configuration.
\begin{figure}[htbp]
    \centering
    \includegraphics[width=1\linewidth]{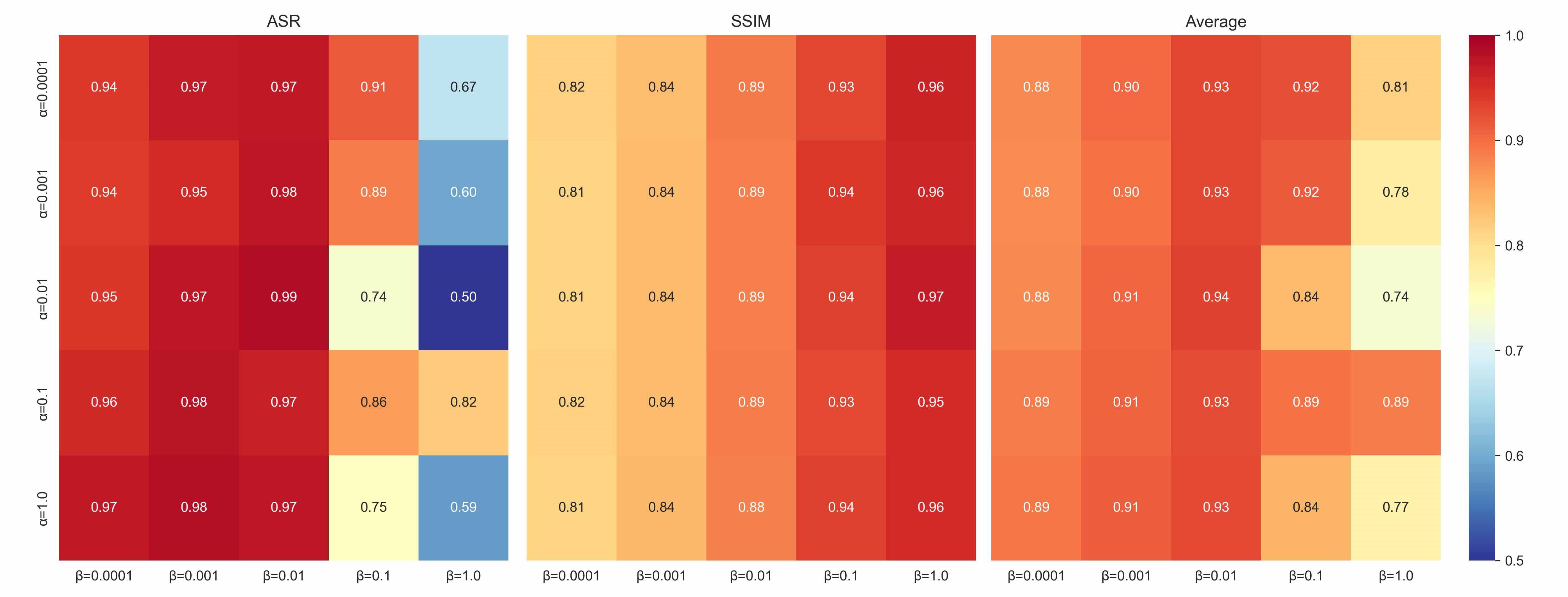}
    \caption{Heatmap of ASR, SSIM, and their average for different configurations of $\alpha$ and $\beta$.}
    \label{fig: score heatmap plot}
\end{figure}

 {We analyze the impact of the NODE system and its time integration step \(N\) on NODE-AdvGAN's performance, focusing on ASR and image similarity measured by PSNR. To explore this, we vary}  \(N\) from 2 to 9 to observe performance changes, as shown in \Cref{fig: N_t plot}. The results show significant performance gains as \(N\) increases to $5$, beyond which further improvements diminish. Notably, our model outperforms the original AdvGAN when \(N\) is set to 3 or higher. {This demonstrates the effectiveness of using the NODE structure as the generator. Based on this analysis, }we select $N=5$ for the optimal value.}

{
After determining the optimal values for \(\alpha\), \(\beta\), and \(N\), we conduct ablation experiments on the noise parameter to {demonstrate the effectiveness of our proposed training strategy and identify an optimal value for maximizing transferability in our NODE-AdvGAN-T method.} For this experiment, we set the test noise level to \(\epsilon = 15/255\). We {systematically} test values of \(\epsilon_\text{train}\) ranging from \(1/255\) to \(15/255\) in increments of $1/255$. {We evaluated the performance of both the original AdvGAN and NODE-AdvGAN models by training them with different \(\epsilon_{\text{train}}\) values and observing the ASR and PSNR values.}
We use VGG16 as the target model during training. Transferability results are performed on VGG19, ResNet18, and DenseNet169. 

The transfer ASR results and PSNR values of adversarial examples for various $\epsilon_{train}$ settings are presented in the left and right columns, respectively, of \Cref{fig: NODE_AdvGAN_changeLinf plot}.
We test the original AdvGAN (top row) and NODE-AdvGAN models (bottom row) and observed that, at approximately \(\epsilon_\text{train} = 10/255\), both models achieve the highest ASR and substantial PSNR values. For both original AdvGAN (top row) and NODE-AdvGAN (bottom row), the highest ASR and competitive PSNR occur at approximately \(\epsilon_\text{train} = 10/255\). {This indicates that our training strategy is not only effective in our proposed model but also improves the performance of the original AdvGAN, demonstrating its generalizability and robustness across different models.} For VGG19, however, better ASR and transferability results are observed when \(\epsilon_\text{train} \geq 10/255\), likely due to architectural similarities with VGG16, which make transfer behavior similar to a white-box attack.  Recognizing that the optimal \(\epsilon_\text{train}\) may not always be feasible in real attack scenarios, we standardized \(\epsilon_\text{train} = 10/255\) across all following NODE-AdvGAN-T implementations. It is worth noting that, ideally, \(\epsilon_\text{train}\) should be adjusted according to the target model in white-box attacks.

\begin{figure}[h]
    \centering
    \begin{minipage}{0.45\textwidth}
        \centering
        \includegraphics[width=\textwidth]{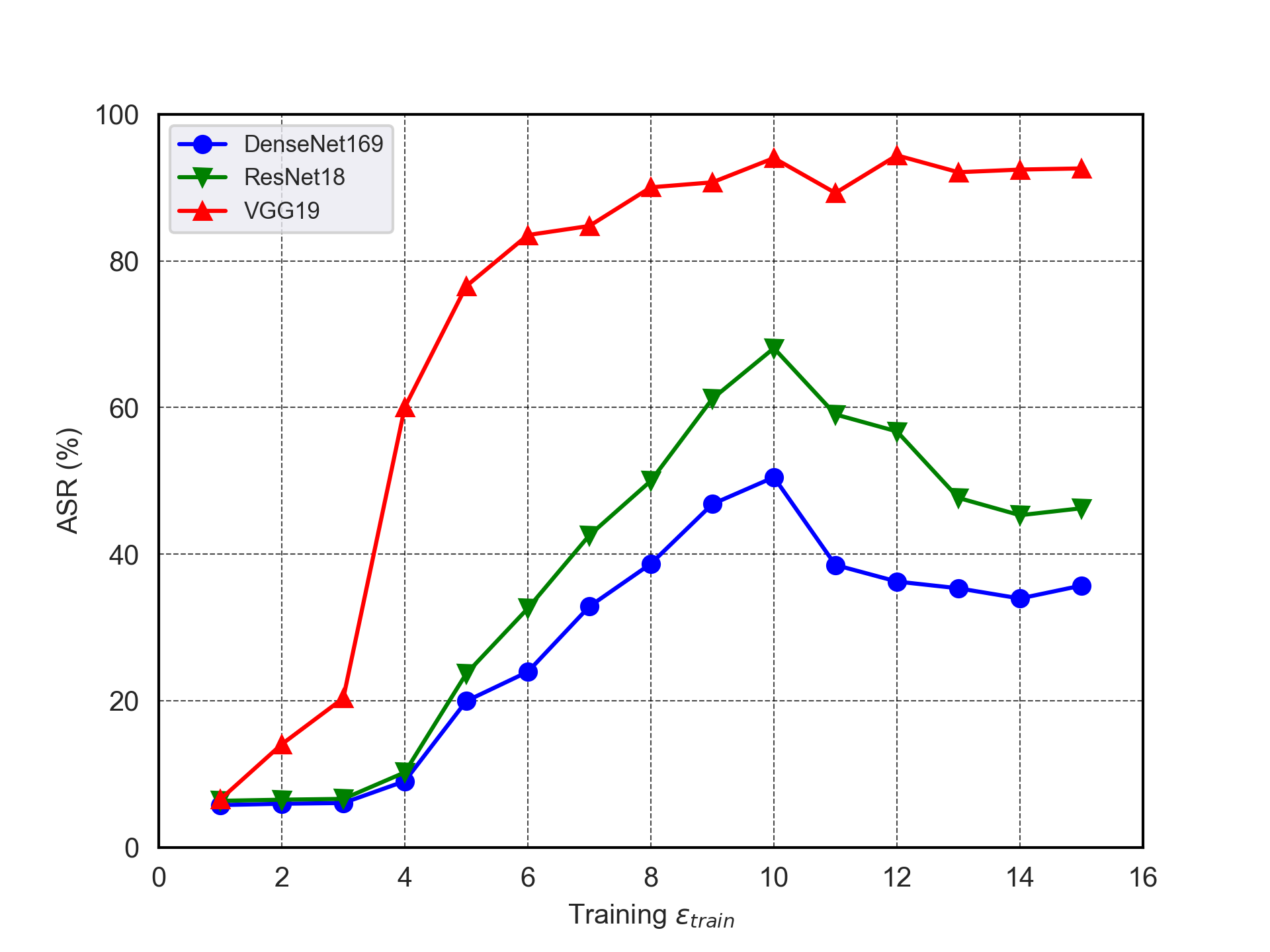} % second image file
    \end{minipage}
    % \hspace{2cm} % specify the exact spacing you want between the images
    \begin{minipage}{0.45\textwidth}
        \centering
        \includegraphics[width=\textwidth]{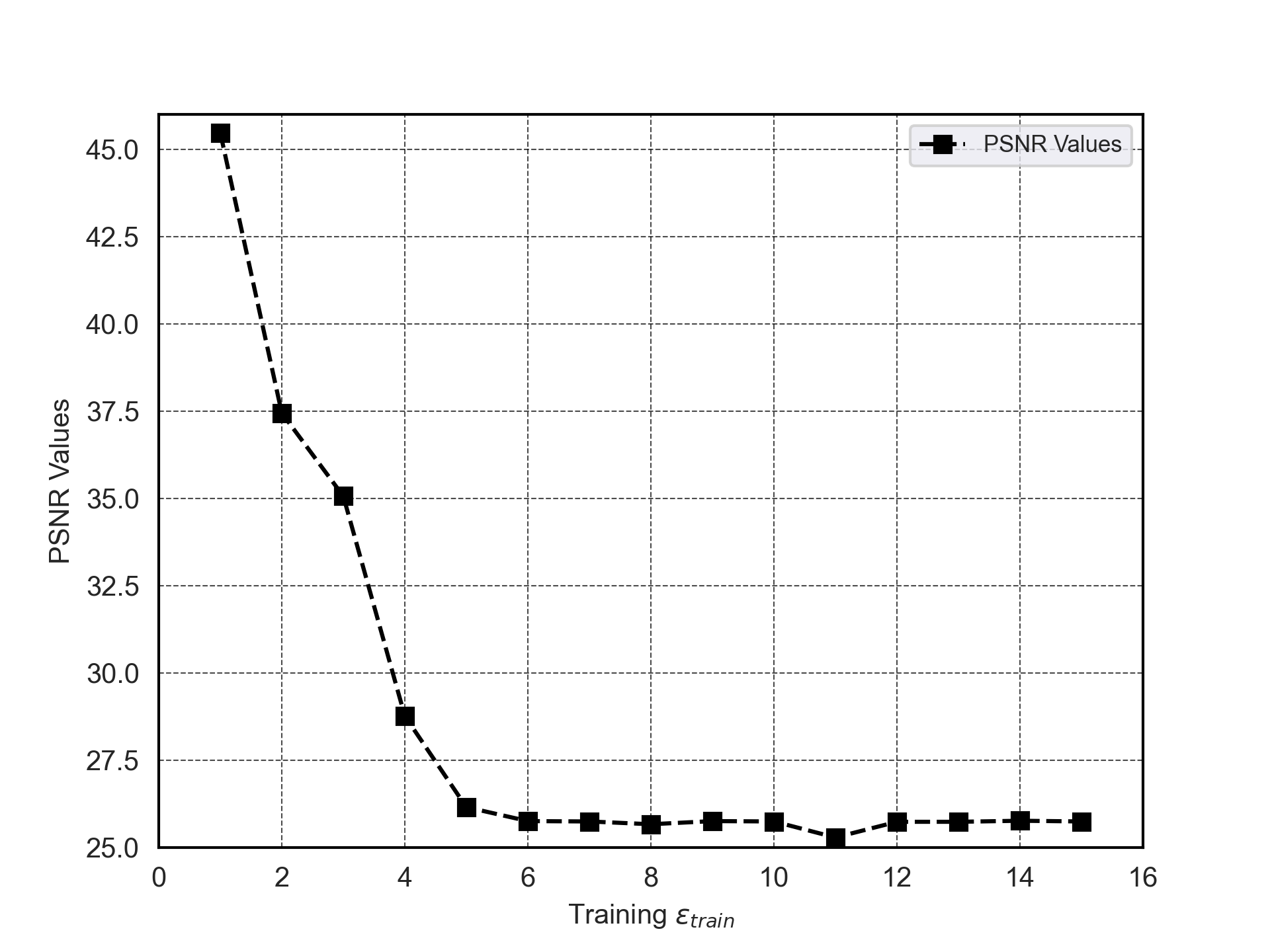} % first image file
    \end{minipage}\\
    \begin{minipage}{0.45\textwidth}
        \centering
        \includegraphics[width=\textwidth]{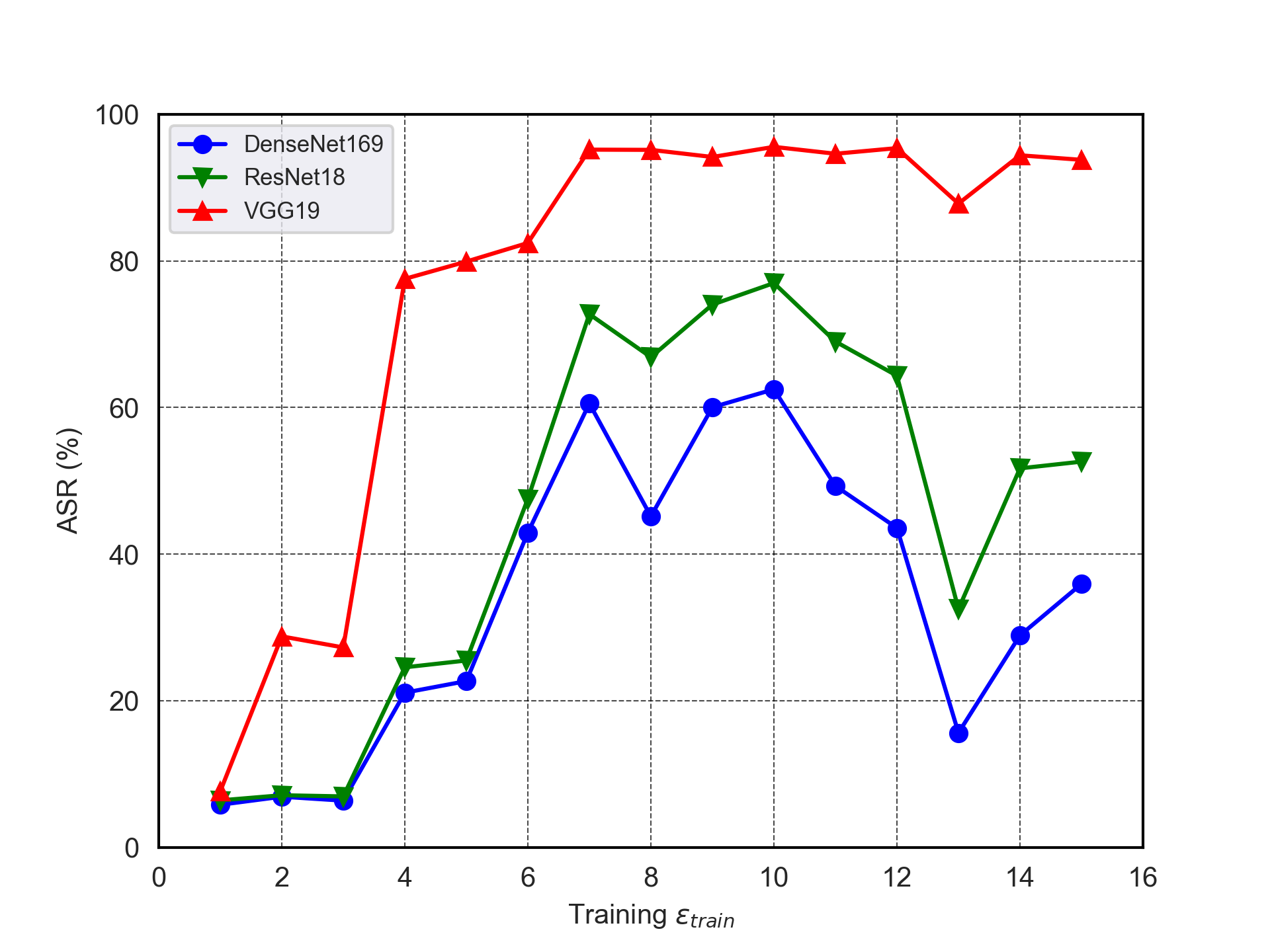} % second image file
    \end{minipage}
    % \hspace{2cm} % specify the exact spacing you want between the images
    \begin{minipage}{0.45\textwidth}
        \centering
        \includegraphics[width=\textwidth]{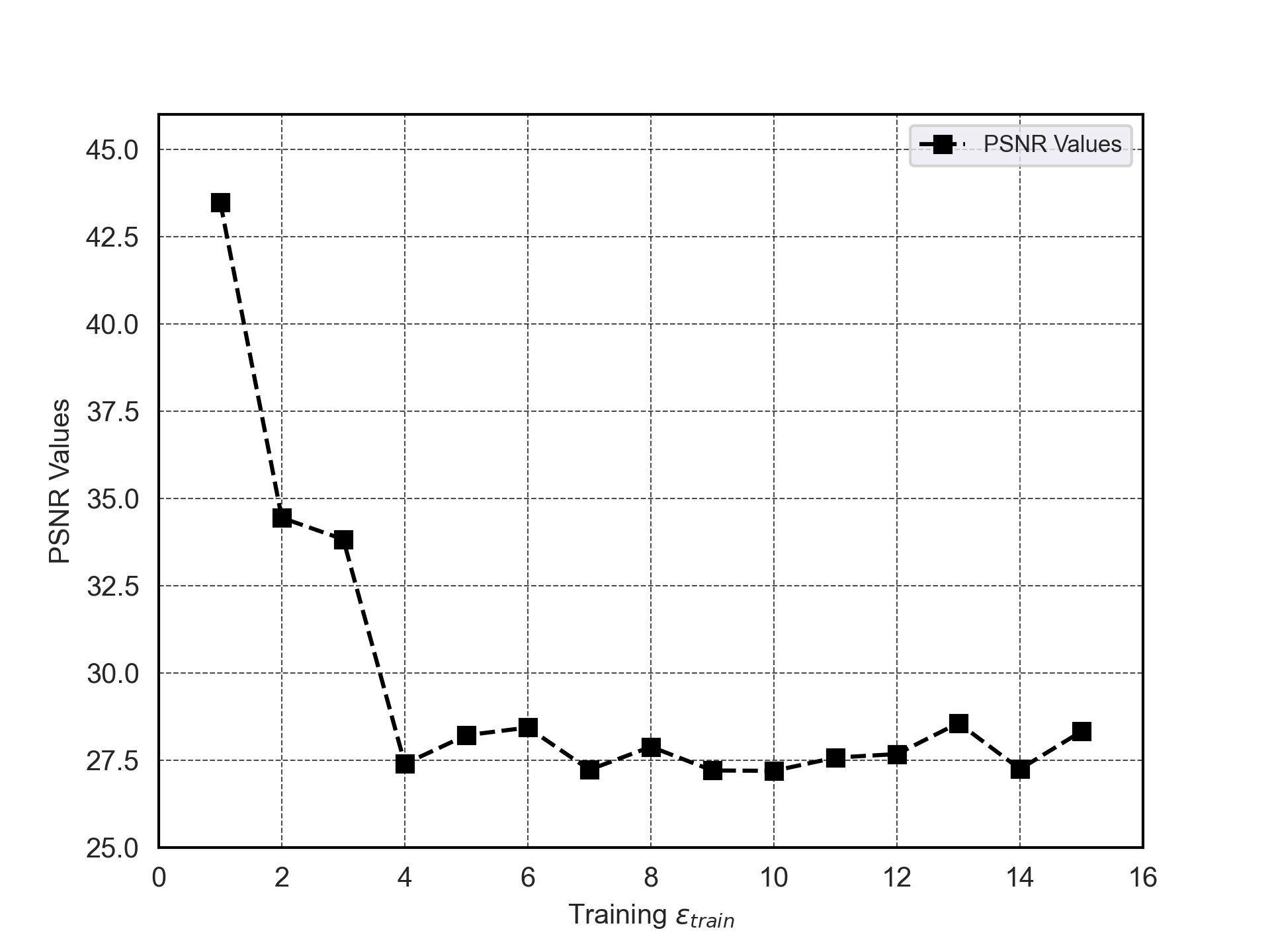} % first image file
    \end{minipage}
    \caption{
    %Comparison of transfererability on 
    Transfer ASR(left) and PSNR(right) for AdvGAN(top) and NODE-AdvGAN-T(bottom) models across  different \(\epsilon_\text{train}\) ($\times 255$) (All experiments evaluated at \(\epsilon=15/255\).).}
    \label{fig: NODE_AdvGAN_changeLinf plot}
\end{figure}

\subsection{Results and Comparisons}
\label{ssec: Results and Comparisons}
\subsubsection{Results on Untargeted Attack}
\label{sssec: Results on Untargeted Attack}
{
In this subsection, we analyze the outcomes of untargeted attacks using our model NODE-AdvGAN on the FMNIST and CIFAR-10 datasets. We examine the performance for white-box (where training and test models for the classification are the same)  and transfer  attacks separately.
For easier representation, we will refer to a transfer attack in which the test target model is different from the training model as a transfer attack. 
For the white-box attacks, we employ Vgg16, ResNet34, and DenseNet121 as target models,  whereas for the transfer attacks, we employ all models described in \Cref{ssec:models}.

For the FMNIST dataset, the results of white-box and transfer attacks are presented in \Cref{tab: FMNIST white-attack} and \Cref{tab: FMNIST transfer attack}, respectively. In the tables, red indicates the highest value for each evaluation metric across methods. From \Cref{tab: FMNIST white-attack}, it is evident that NODE-AdvGAN achieves the highest ASR in white-box attacks while maintaining relatively high SSIM and PSNR for all attacked models. Compared to other gradient-based noise generation methods, our model shows significant improvement and consistently outperforms the original AdvGAN across all evaluated metrics. Finally, we observed similar performance between NODE-AdvGAN and NODE-AdvGAN-T in white-box attacks.

In \Cref{tab: FMNIST transfer attack} of the transfer attacks, our NODE-AdvGAN-T ($\epsilon_\text{train}=10/255$) model consistently yields the best performance across all listed experiments.  We again observe significantly better transfer results than gradient-based methods and the original AdvGAN. Moreover, thanks to the tuning of training noise parameter $\epsilon_\text{train}$,  the NODE-AdvGAN-T model also shows  notable improvements {in ASR for transfer attacks,} ranging from $9.43\%$ to $29.89\%$, with an average improvement of $17.96\%$ than the NODE-AdvGAN model. 

%Simultaneously, our NODE-AdvGAN method also exhibits superior performance compared to the original AdvGAN model, outperforming it in nearly all the transfer attacks except for a few.

\begin{table}[htbp]
\caption{
White-box untargeted attack on the FMNIST dataset.
%Untargeted attack success rate (ASR: \%) under white-box settings on the FMNIST dataset.
}
\label{tab: FMNIST white-attack}
\resizebox{\textwidth}{!}{%
\begin{tabular}{@{}llllllllll@{}}
\toprule
                         & \multicolumn{3}{l}{VGG16}                                                                     & \multicolumn{3}{l}{ResNet34}                                                                  & \multicolumn{3}{l}{DenseNet121}                                                               \\ 
\cmidrule(lr){2-4} \cmidrule(lr){5-7} \cmidrule(lr){8-10}
\multirow{-2}{*}{} & ASR                             & SSIM                          & PSNR                         & ASR                             & SSIM                          & PSNR                         &ASR                             & SSIM                          & PSNR                         \\ \midrule
FGSM                     & 30.58\%                        & 0.7529                        & 25.87                        & 37.85\%                        & 0.7543                        & 25.89                        & 43.92\%                        & 0.7539                        & 25.86                        \\
I-FGSM                      & 91.39\%                        & {\color[HTML]{9C0006} \textbf{0.8739}} & {\color[HTML]{9C0006} \textbf{30.91}} & 88.46\%                        & {\color[HTML]{9C0006} \textbf{0.8757}} & {\color[HTML]{9C0006} \textbf{31.01}} & 91.52\%                        & {\color[HTML]{9C0006} \textbf{0.8782}} & {\color[HTML]{9C0006} \textbf{31.52}} \\
MI-FGSM                  & 83.55\%                        & 0.7971                        & 27.52                        & 79.71\%                        & { 0.7989}                     & {27.48}                      & 90.59\%                        & 0.7864                        & 27.16                        \\
Ni-FGSM                  & 92.93\%                        & 0.7880                        & 27.40                        & 92.16\%                        & 0.7910                        & 27.40                        & 88.08\%                        & 0.7821                        & 27.11                        \\
AdvGAN                   & 98.66\%                        & 0.7829                        & 27.34                        & 97.14\%                        & 0.7793                        & 26.68                        & 97.74\%                        & 0.7852                        & 27.23                        \\
NODE-AdvGAN                     & {\color[HTML]{9C0006}\textbf{99.10\%}}                        & {0.8060}                      & {28.19}                      & {\color[HTML]{9C0006} \textbf{97.94\%}}& 0.7797                        & 26.86                        & {\color[HTML]{9C0006} \textbf{98.24\%}} & {0.8011}                      & {27.51}                      \\
NODE-AdvGAN-T        & {99.06\%} & 0.7813                        & 27.16                        &93.19\%                         & 0.7897                        & 26.52                        & 97.21\%                        & 0.7810                        & 27.11                        \\
\bottomrule
\end{tabular}%
}
\end{table}

\begin{table}[htbp]
\caption{
ASR of transfer untargeted attack on the FMNIST dataset (Results marked with * are from white-box attacks).
}
\label{tab: FMNIST transfer attack}
\resizebox{\textwidth}{!}{%
\begin{tabular}{@{}clllllll@{}}
\toprule
\multicolumn{1}{l}{}          & Attack                      & VGG16                          & VGG19                          & ResNet34                       & ResNet18                       & DenseNet121                    & DenseNet169                     \\ \midrule
                              & FGSM    & 30.61\%                        & 38.56\%                        & 37.39\%                        & 42.39\%                        & *43.92\%                         & 37.80\%                        \\
                              & I-FGSM     & 22.53\%                        & 21.94\%                        & 25.30\%                        & 26.92\%                        & *91.52\%                       & 32.83\%                        \\
                              & MI-FGSM & 33.28\%                        & 36.83\%                        & 41.11\%                        & 46.85\%                        & *90.59\%                        & 50.26\%                        \\
                              & Ni-FGSM & 34.12\%                        & 38.93\%                        & 42.02\%                        & 47.62\%                        & *88.08\%                         & 51.77\%                        \\
                              & AdvGAN  & 28.87\%                        & 33.59\%                        & 30.96\%                        & 38.70\%                        & *97.74\%                        & 59.97\%                        \\
                              & NODE-AdvGAN & 32.46\%                    & 33.32\%                        & 33.21\%                        & 39.05\%                        & *98.24\%                      & 62.75\%                        \\
\multirow{-7}{*}{DenseNet121} & NODE-AdvGAN-T & {\color[HTML]{9C0006}\textbf{45.56\%} } & {\color[HTML]{9C0006} \textbf{47.54\%}} & {\color[HTML]{9C0006} \textbf{52.16\%}} & {\color[HTML]{9C0006} \textbf{60.31\%}} &*97.21\%                  & {\color[HTML]{9C0006} \textbf{84.70\%}} \\
                              &         &                                &                                &                                &                                &                                &                                \\
                              & FGSM    & 28.34\%                        & 35.21\%                        &*37.85\%                         & 39.96\%                        & 33.77\%                        & 31.98\%                        \\
                              & I-FGSM     & 28.64\%                        & 29.39\%                        & *88.46\%                         & 40.32\%                        & 36.93\%                        & 33.01\%                        \\
                              & MI-FGSM & 34.90\%                        & 37.96\%                        &*79.71\%                       & 48.16\%                        & 42.71\%                        & 39.67\%                        \\
                              & Ni-FGSM & 38.59\%                        & 41.95\%                        &  *92.16\%                        & 53.65\%                        & 47.49\%                        & 43.93\%                        \\
                              & AdvGAN  & 39.96\%                        & 55.97\%                        &  *97.14\%                         & 72.43\%                        & 69.99\%                        & 50.89\%                        \\
                              & NODE-AdvGAN    & 41.57\%                        & 57.95\%                 &  *97.94\%                        & 75.11\%                        & 70.50\%                        & 50.78\%                        \\
\multirow{-7}{*}{ResNet34}    & NODE-AdvGAN-T    & {\color[HTML]{9C0006} \textbf{71.46\%}} & {\color[HTML]{9C0006} \textbf{74.85\%}} & *93.19\%          & {\color[HTML]{9C0006} \textbf{85.86\%}} & {\color[HTML]{9C0006} \textbf{82.33\%}} & {\color[HTML]{9C0006} \textbf{73.43\%}} \\
                              &         &                         &                                &                                &                                &                                &                                \\
                              & FGSM    &*30.58\%                        & 35.63\%                        & 30.32\%                        & 34.98\%                        & 29.94\%                        & 29.90\%                        \\
                              & I-FGSM     &*91.39\%                        & 40.44\%                        & 23.70\%                        & 24.44\%                        & 24.58\%                        & 22.06\%                        \\
                              & MI-FGSM & *83.55\%                       & 47.92\%                        & 33.85\%                        & 36.97\%                        & 33.56\%                        & 33.18\%                        \\
                              & Ni-FGSM &  *92.93\%                      & 51.98\%                        & 35.19\%                        & 40.82\%                        & 36.74\%                        & 35.30\%                        \\
                              & AdvGAN  &  *98.66\%                      & 73.18\%                        & 29.23\%                        & 34.17\%                        & 30.52\%                        & 33.12\%                        \\
                              & NODE-AdvGAN    & *99.10\%                         & 68.74\%                        & 29.27\%                        & 34.39\%                        & 34.19\%                        & 33.94\%                        \\
\multirow{-7}{*}{VGG16}       & NODE-AdvGAN-T    &*99.06\%                       & {\color[HTML]{9C0006} \textbf{94.56\%}} & {\color[HTML]{9C0006} \textbf{47.81\%}} & {\color[HTML]{9C0006} \textbf{56.44\%}} & {\color[HTML]{9C0006} \textbf{46.32\%}} & {\color[HTML]{9C0006} \textbf{43.37\%}} \\ \bottomrule
\end{tabular}%
}
\end{table}
% \footnotetext{Results marked with * are from white-box attacks.} 

For the CIFAR-10 dataset, 
 the results of the white-box and transfer attacks are shown in \Cref{tab: CIFAR-10 white-attack} and  \Cref{tab: CIFAR-10 transfer attack}, respectively. We observe similar results as for FMNIST data sets.

\begin{table}[htbp]
\caption{white-box untargeted attack on the  CIFAR-10 dataset.}
\label{tab: CIFAR-10 white-attack}
\resizebox{\textwidth}{!}{%
\begin{tabular}{@{}llllllllll@{}}
\toprule
                         & \multicolumn{3}{c}{VGG16}                                         & \multicolumn{3}{c}{ResNet34}                                      & \multicolumn{3}{c}{DenseNet121}                                   \\
\cmidrule(lr){2-4} \cmidrule(lr){5-7} \cmidrule(lr){8-10}
                         & ASR            & SSIM      & PSNR         & ASR            & SSIM      & PSNR         & ASR            & SSIM      & PSNR         \\ \midrule
FGSM                     & 54.27\%       & 0.8050    & 24.73        & 56.01\%       & 0.8081    & 24.73        & 39.86\%       & 0.8100    & 24.73        \\
I-FGSM                      & 90.53\%       & {\color[HTML]{9C0006}\textbf{0.9420}}    & {\color[HTML]{9C0006}\textbf{31.38}}        & 91.81\%       & {\color[HTML]{9C0006}\textbf{0.9402}}    & {\color[HTML]{9C0006}\textbf{30.96}}        & 69.09\%       & {\color[HTML]{9C0006}\textbf{0.9351}}    & {\color[HTML]{9C0006}\textbf{30.56}}        \\
MI-FGSM                  & 81.90\%       & 0.8562    & 26.58        & 85.88\%       & 0.8584    & 26.55        & 63.24\%       & 0.8589    & 26.49        \\
Ni-FGSM                  & 90.31\%       & 0.8501    & 26.42        & {\color[HTML]{9C0006}\textbf{97.48\%}}       & 0.8556    & 26.51        & 72.66\%       & 0.8573    & 26.49        \\
AdvGAN                   & 97.06\%       & 0.8240    & 25.74        & 86.87\%       & 0.8361    & 25.80        & 74.67\%       & 0.8110    & 24.82        \\
NODE-AdvGAN                     & {\color[HTML]{9C0006}\textbf{97.54\%}}       & 0.8610    & 27.27        & 89.39\%       & 0.8718    & 26.76        & 87.09\%       & 0.8671    & 26.74        \\
NODE-AdvGAN-T                     & 96.51\%       & 0.8649    & 27.19        & 88.59\%       & 0.8689    & 26.72        & {\color[HTML]{9C0006}\textbf{88.52\%}}       & 0.8804    & 26.97        \\
\bottomrule
\end{tabular}%
}
\end{table}

\begin{table}[htbp]
\caption{ASR of transfer untargeted attack on the CIFAR-10 dataset (Results marked with * are from white-box attacks).}
\label{tab: CIFAR-10 transfer attack}
\resizebox{\textwidth}{!}{%
\begin{tabular}{@{}clllllll@{}}
\toprule
\multicolumn{1}{l}{}          & Attack                      & VGG16                          & VGG19                          & ResNet34                       & ResNet18                       & DenseNet121                    & DenseNet169                    \\ \midrule
                              & FGSM                        & 45.77\%                        & 46.32\%                        & 39.10\%                        & 38.31\%                        & *39.86\%                        & 39.30\%                        \\
                              & I-FGSM                         & 45.15\%                        & 42.97\%                        & 45.73\%                        & 46.81\%                        & *69.09\%                       & 57.08\%                        \\
                              & MI-FGSM                     & 51.98\%                        & 50.77\%                        & 54.16\%                        & 54.57\%                        & *63.24\%                        & 58.50\%                        \\
                              & Ni-FGSM                     & 58.84\%                        & 57.33\%                        & 58.72\%                        & 59.36\%                        & *72.66\%                        & 65.62\%                        \\
                              & AdvGAN                      & 71.97\%                        & 67.69\%                        & 68.62\%                        & 67.90\%                        & *74.67\%                       & 77.50\%                        \\
                              & NODE-AdvGAN                        & 84.98\%                        & 84.34\%                        & 71.88\%                        & 68.57\%                 & *87.09\%                       & 80.47\%                        \\
\multirow{-7}{*}{DenseNet121} & NODE-AdvGAN-T                        & {\color[HTML]{9C0006} \textbf{87.63\%}} & {\color[HTML]{9C0006} \textbf{87.91\%}} & {\color[HTML]{9C0006} \textbf{82.54\%}} & {\color[HTML]{9C0006} \textbf{82.63\%}} & *88.52\%               & {\color[HTML]{9C0006} \textbf{87.46\%}} \\
\multicolumn{1}{l}{}          &                             &                                &                                &                                &                                &                                &                                \\
                              & FGSM                        & 61.49\%                        & 61.01\%                        & *56.01\%                        & 51.42\%                        & 50.31\%                        & 51.36\%                        \\
                              & I-FGSM                         & 68.95\%                        & 65.70\%                        & *91.81\%                        & 75.48\%                        & 61.69\%                        & 51.36\%                        \\
                              & MI-FGSM                     & 74.95\%                        & 73.36\%                        & *85.88\%                        & 78.27\%                        & 73.12\%                        & 72.88\%                        \\
                              & Ni-FGSM                     & 85.87\%                        & 83.74\%                        & *97.48\%                        & {\color[HTML]{9C0006}\textbf{88.42\%}}  & 81.76\%                        & 81.79\%                        \\
                              & AdvGAN                      & 86.93\%                        & 87.57\%                        & *86.87\%                        & 84.75\%                        & 83.85\%                        & 81.14\%                        \\
                              & NODE-AdvGAN                        & 87.90\%                        & 89.04\%                 & *89.39\%                       & 86.14\%                        & 84.31\%                        & 85.30\%                        \\
\multirow{-7}{*}{ResNet34}    & NODE-AdvGAN-T                        & {\color[HTML]{9C0006} \textbf{88.89\%}} & {\color[HTML]{9C0006} \textbf{89.20\%}} & *88.59\%                &  87.90\%                       & {\color[HTML]{9C0006} \textbf{87.86\%}} & {\color[HTML]{9C0006} \textbf{87.28\%}} \\
\multicolumn{1}{l}{}          &                             &                                 &                                &                                &                                &                                &                                \\
                              & FGSM                        & *54.27\%                          & 56.73\%                        & 43.16\%                        & 43.23\%                        & 41.62\%                        & 41.70\%                        \\
                              & I-FGSM                         & *90.53\%                         & 68.29\%                        & 32.78\%                        & 34.67\%                        & 27.57\%                        & 26.86\%                        \\
                              & MI-FGSM                     & *81.90\%                        & 71.44\%                        & 57.67\%                        & 57.62\%                        & 51.97\%                        & 52.22\%                        \\
                              & Ni-FGSM                     & *90.31\%                         & 74.68\%                        & 54.99\%                        & 55.21\%                        & 50.11\%                        & 50.09\%                        \\
                              & AdvGAN                      &*97.54\%                        & 92.61\%                        & 45.58\%                        & 46.25\%                        & 34.68\%                        & 35.70\%                        \\
                              & {NODE-AdvGAN}               &*97.54\%                        & 93.78\%                        & 45.59\%                        & 52.61\%                        & 40.60\%                        & 35.93\%                        \\
\multirow{-7}{*}{VGG16}       & NODE-AdvGAN-T               &*96.51\%                         & {\color[HTML]{9C0006} \textbf{95.55\%}} & {\color[HTML]{9C0006} \textbf{71.15\%}} & {\color[HTML]{9C0006} \textbf{76.97\%}} & {\color[HTML]{9C0006} \textbf{65.90\%}} & {\color[HTML]{9C0006} \textbf{62.48\%}} \\ 
\bottomrule
\end{tabular}%
}
\end{table}

In summary, experiments across both datasets demonstrate that our models generate highly effective adversarial images with robust attack capabilities in transfer and white-box untargeted attacks, applicable to both greyscale and colour images. Specifically, our models show significant improvements over all gradient-based noise generation methods across attack scenarios and consistently outperform the original AdvGAN, particularly in transfer attacks, due to optimized noise parameter tuning. In white-box attacks, NODE-AdvGAN performs similarly to NODE-AdvGAN-T, but NODE-AdvGAN-T achieves notably better transfer results.

We also compare the computation/time cost on DenseNet121 as the target model on the CIFAR-10 dataset. 
A run time is measured for generating 10,000 adversarial instances during test time. The outcomes are displayed in \Cref{tab: generation time}. As observed, our model outperforms other models in speed, except for being slower than AdvGAN. However, it is essential to note that generator-based models require training. 

\begin{table}[htbp]
\caption{
Comparison on time efficiency with attack model DenseNet121.
%The generation efficiency of adversarial samples on the CIFAR-10 dataset
}
\label{tab: generation time}
\resizebox{\textwidth}{!}{%
{\scriptsize
\begin{tabular}{llllllll}
\hline
Attack & FGSM  & I-FGSM    & MI-FGSM & NI-FGSM & AdvGAN         & NODE-AdvGAN  & NODE-AdvGAN-T  \\ \hline
Time   & 2.60s & 17.10s & 17.19s  & 17.28s  & \textbf{0.12s} & 1.08s & 1.08s \\ \hline
\end{tabular}%
}}
\end{table}

\subsubsection{Results on Targeted Attack}
\label{sssec: Results on Targeted Attack}
In adversarial machine learning, targeted attacks are much more difficult than untargeted attacks. This is mainly because the optimization of targeted attack navigates a more complex and narrow path to alter the model's decision specifically toward the target class. This involves finding a perturbation that not only pushes the input out of its current class but also precisely into the target class, resulting in a more constrained and complex optimization problem.

Given the multiple categories, we present experiments using DenseNet121 as the white-box target model on the CIFAR-10 dataset for targeted attacks. The results are shown in \cref{tab: targeted white-attacks}. Our findings indicate that our method and AdvGAN significantly outperform gradient-based methods in white-box testing. {This discrepancy may stem from two factors: (1) the inherently constrained optimization landscapes of targeted attacks compared to untargeted scenarios  \cite{carlini2017towards}, and (2) the model’s training-time data augmentation strategies (e.g., random cropping), which are known to disrupt gradient alignment between inputs and adversarial targets \cite{xie2019feature}.}

Compared to the original AdvGAN, our models NODE-AdvGAN and NODE-AdvGAN-T achieved significantly higher targeted ASR of $96.86\%$ and $96.37\%$, respectively, surpassing AdvGAN's $85.48\%$. Additionally, NODE-AdvGAN and NODE-AdvGAN-T demonstrated greater similarity to the original images, with PSNR values higher than the original AdvGAN by $2.51,\text{dB}$ and $1.71,\text{dB}$, respectively. This indicates that our methods maintain high ASR and introduce less perturbation.

\begin{table}[htbp]
\caption{
White-box targeted attack on the CIFAR-10 dataset.
}
\label{tab: targeted white-attacks}
\resizebox{\textwidth}{!}{%
\begin{tabular}{@{}lllllllllllll@{}}
\toprule 
Target Class   & \multicolumn{2}{l}{I-FGSM}       & \multicolumn{2}{l}{MI-FGSM} & \multicolumn{2}{l}{NI-FGSM}   & \multicolumn{2}{l}{AdvGAN}   & \multicolumn{2}{l}{NODE-AdvGAN}   & \multicolumn{2}{l}{NODE-AdvGAN-T}     \\ \cmidrule(l){2-3} \cmidrule(l){4-5} \cmidrule(l){6-7} \cmidrule(l){8-9} \cmidrule(l){10-11} \cmidrule(l){12-13} 
        & ASR    & PSNR                         & ASR     & PSNR  & ASR     & PSNR  & ASR     & PSNR  & ASR                           & PSNR  & ASR                           & PSNR  \\ \midrule
0       & 13.04 & {\color[HTML]{9C0006} \textbf{30.92}} & 16.60  & 26.48 & 16.56  & 26.49 & 89.81  & 25.16 & {\color[HTML]{9C0006} \textbf{95.62}} & 27.44 & 93.39                        & 27.01 \\
1       & 10.23 & {\color[HTML]{9C0006} \textbf{31.05}} & 16.17  & 26.50 & 18.57  & 26.53 & 87.34  & 25.30 & {\color[HTML]{9C0006} \textbf{91.90}} & 27.42 & 89.90                        & 26.58 \\
2       & 16.26 & {\color[HTML]{9C0006} \textbf{30.85}} & 25.20  & 26.50 & 28.24  & 26.51 & 86.46  & 25.26 & 97.61                        & 28.02 & {\color[HTML]{9C0006} \textbf{98.23}} & 26.56 \\
3       & 19.00 & {\color[HTML]{9C0006} \textbf{30.77}} & 23.33  & 26.50 & 23.52  & 26.51 & 96.86  & 25.89 & 97.71                        & 27.66 & {\color[HTML]{9C0006} \textbf{98.58}} & 26.97 \\
4       & 17.73 & {\color[HTML]{9C0006} \textbf{30.86}} & 22.22  & 26.53 & 24.24  & 26.54 & 97.00  & 25.17 & 98.66                        & 28.65 & {\color[HTML]{9C0006} \textbf{98.81}} & 27.21 \\
5       & 16.08 & {\color[HTML]{9C0006} \textbf{30.82}} & 21.58  & 26.51 & 22.54  & 26.53 & 91.79  & 25.55 & {\color[HTML]{9C0006} \textbf{93.88}} & 26.76 & 91.37                        & 27.37 \\
6       & 18.88 & {\color[HTML]{9C0006} \textbf{30.59}} & 23.98  & 26.55 & 30.80  & 26.58 & 73.13  & 25.11 & 98.93                        & 28.63 & {\color[HTML]{9C0006} \textbf{99.51}} & 26.68 \\
7       & 10.43 & {\color[HTML]{9C0006} \textbf{30.96}} & 18.31  & 26.52 & 19.61  & 26.55 & 88.32  & 25.09 & {\color[HTML]{9C0006} \textbf{97.80}} & 28.04 & 96.53                        & 27.07 \\
8       & 9.46  & {\color[HTML]{9C0006} \textbf{30.87}} & 12.14  & 26.52 & 12.47  & 26.53 & 69.49  & 25.13 & 98.36                        & 27.60 & {\color[HTML]{9C0006} \textbf{98.52}} & 27.19 \\
9       & 14.33 & {\color[HTML]{9C0006} \textbf{31.10}} & 20.90  & 26.49 & 24.03  & 26.51 & 74.63  & 25.16 & 98.17                        & 27.68 & {\color[HTML]{9C0006} \textbf{98.87}} & 27.21 \\
Average & 14.54 & {\color[HTML]{9C0006} \textbf{30.88}} & 20.04  & 26.51 & 22.06  & 26.53 & 85.48  & 25.28 & {\color[HTML]{9C0006} \textbf{96.86}} & 27.79 & 96.37                        & 26.99 \\ \bottomrule
\end{tabular}%
}
\end{table}

The results of transfer attacks using AdvGAN, NODE-AdvGAN, and NODE-AdvGAN-T are presented in \Cref{tab: targeted transfer attacks}. Given the poor performance of gradient-based attacks, we have omitted those comparisons. Similar to the untargeted attack results, we observe that both NODE-AdvGAN and NODE-AdvGAN-T exhibit superior transfer attack capabilities compared to AdvGAN. Additionally, the NODE-AdvGAN-T method further enhances transfer attack capability beyond that of NODE-AdvGAN.

\begin{table}[htbp]
\caption{ASR of transfer Targeted Attacks Transferred from DenseNet121 on CIFAR-10 Dataset. In this table, ``NODE-A'' refers to the NODE-AdvGAN model, and ``NODE-A-T'' refers to the NODE-AdvGAN-T model.}
\label{tab: targeted transfer attacks}
\resizebox{\textwidth}{!}{%
\begin{tabular}{@{}llllllllll@{}}
\toprule
Class   & \multicolumn{3}{l}{DenseNet169}                                                            & \multicolumn{3}{l}{ResNet18}                                                               & \multicolumn{3}{l}{VGG19}                                                                  \\ \cmidrule(lr){2-4} \cmidrule(lr){5-7} \cmidrule(lr){8-10}
        & AdvGAN                       & NODE-A                         & NODE-A-T                         & AdvGAN                       & NODE-A                         & NODE-A-T                         & AdvGAN                       & NODE-A                        & NODE-A-T                         \\ \midrule
0       & 81.22\%                        & {\color[HTML]{9C0006} \textbf{93.69\%}} & 93.61\%                        & 78.53\%                        & {\color[HTML]{9C0006} \textbf{84.64\%}} & 82.66\%                        & {\color[HTML]{9C0006} \textbf{92.47\%}} & 87.19\%                        & 85.72\%                        \\
1       & 82.53\%                        & 74.89\%                        & {\color[HTML]{9C0006} \textbf{90.21\%}} & 81.43\%                        & 61.41\%                        & {\color[HTML]{9C0006} \textbf{83.23\%}} & 88.94\%                        & 77.74\%                        & {\color[HTML]{9C0006} \textbf{91.23\%}} \\
2       & 72.76\%                        & 69.71\%                        & {\color[HTML]{9C0006} \textbf{93.84\%}} & 74.34\%                        & 40.81\%                        & {\color[HTML]{9C0006} \textbf{82.53\%}} & 77.64\%                        & 45.38\%                        & {\color[HTML]{9C0006} \textbf{87.51\%}} \\
3       & 92.71\%                        & 95.34\%                        & {\color[HTML]{9C0006} \textbf{96.71\%}} & 90.88\%                        & 87.00\%                        & {\color[HTML]{9C0006} \textbf{93.58\%}} & 96.44\%                        & 95.21\%                        & {\color[HTML]{9C0006} \textbf{97.07\%}} \\
4       & 95.74\%                        & 97.46\%                        & {\color[HTML]{9C0006} \textbf{97.54\%}} & 92.53\%                        & 95.26\%                        & {\color[HTML]{9C0006} \textbf{96.41\%}} & 91.23\%                        & 96.98\%                        & {\color[HTML]{9C0006} \textbf{97.39\%}} \\
5       & {\color[HTML]{9C0006} \textbf{88.82\%}} & 88.76\%                        & 85.19\%                        & {\color[HTML]{9C0006} \textbf{82.32\%}} & 77.26\%                        & 80.48\%                        & {\color[HTML]{9C0006} \textbf{89.10\%}} & 78.12\%                        & 78.36\%                        \\
6       & 54.44\%                        & 96.63\%                        & {\color[HTML]{9C0006} \textbf{98.21\%}} & 9.17\%                         & 94.10\%                        & {\color[HTML]{9C0006} \textbf{96.90\%}} & 39.21\%                        & 97.88\%                        & {\color[HTML]{9C0006} \textbf{98.74\%}} \\
7       & 72.33\%                        & {\color[HTML]{9C0006} \textbf{95.93\%}} & 92.76\%                        & 57.12\%                        & {\color[HTML]{9C0006} \textbf{94.01\%}} & 91.70\%                        & 33.56\%                        & {\color[HTML]{9C0006} \textbf{93.82\%}} & 88.27\%                        \\
8       & 51.56\%                        & 96.74\%                        & {\color[HTML]{9C0006} \textbf{97.71\%}} & 35.48\%                        & 93.89\%                        & {\color[HTML]{9C0006} \textbf{95.09\%}} & 50.44\%                        & 96.20\%                        & {\color[HTML]{9C0006} \textbf{95.52\%}} \\
9       & 59.47\%                        & 94.08\%                        & {\color[HTML]{9C0006} \textbf{96.83\%}} & 48.04\%                        & 92.78\%                        & {\color[HTML]{9C0006} \textbf{95.40\%}} & 53.47\%                        & 96.58\%                        & {\color[HTML]{9C0006} \textbf{97.11\%}} \\
Average & 75.16\%                        & 90.32\%                        & {\color[HTML]{9C0006} \textbf{94.26\%}} & 64.99\%                        & 82.12\%                        & {\color[HTML]{9C0006} \textbf{89.80\%}} & 71.25\%                        & 86.51\%                        & {\color[HTML]{9C0006} \textbf{91.69\%}} \\ \bottomrule
\end{tabular}%
}
\end{table}

\Cref{pic: show Samples} provides a visualization of adversarial examples generated by NODE-AdvGAN (left) and NODE-AdvGAN-T (right) on the CIFAR-10 dataset. The images along the diagonal are clean inputs, while the off-diagonal images are adversarial examples. Specifically, in the left (right) figure, the image in the \(i\)-th row and \(j\)-th column (\(i \neq j\)) is an adversarial example generated by NODE-AdvGAN (NODE-AdvGAN-T), using the clean image in the \(i\)-th row and \(i\)-th column as input and targeting the \(j\)-th category. 

\begin{figure}[htbp]
    \centering
    \begin{minipage}{0.45\textwidth}
        \centering
        \includegraphics[width=\textwidth]{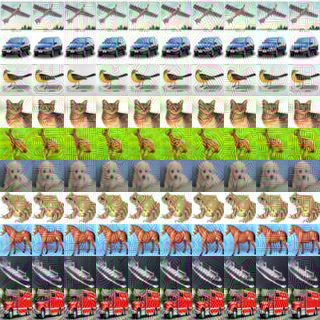} % second image file
    \end{minipage}
    % \hspace{2cm} % specify the exact spacing you want between the images
    \begin{minipage}{0.45\textwidth}
        \centering
        \includegraphics[width=\textwidth]{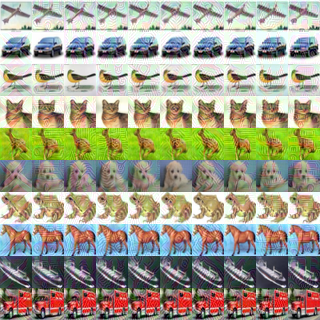} % first image file
    \end{minipage}
    \caption{Adversarial samples generated from targeted attacks by NODE-AdvGAN (left) and NODE-AdvGAN-T (right).}
    \label{pic: show Samples}
\end{figure}

%% file: conclusion.tex
\section{Conclusion}
\label{sec: Conclusion}
{
In this paper, we presented NODE-AdvGAN to improve the quality and transferability of adversarial examples. By leveraging a dynamic-system perspective, we treated the adversarial generation as a continuous process, mimicking the iterative nature of traditional gradient-based methods. This approach allowed us to generate smoother and more precise perturbations while preserving important features of the original images. We also introduced NODE-AdvGAN-T, which enhances transferability by tuning the noise parameters during training.

Our experiments showed that NODE-AdvGAN and NODE-AdvGAN-T achieve higher attack success rates and better image quality than existing methods. This work demonstrates the effectiveness of combining dynamic systems with machine learning to address adversarial robustness in AI models.

For future work, we could focus on defending against adversarial attacks. One approach would be using our generated adversarial examples to train classifier models, making them more robust against such attacks. We could also explore combining multiple classification models to create a super-classifier with superior accuracy and robustness. Additionally, integrating our approach with other generator structures, such as diffusion models, could further enhance adversarial generation. Finally, we could investigate more efficient NODE architectures or other continuous-time dynamic models to reduce computational costs while maintaining effectiveness.}